%% file: main.tex
\documentclass[journal]{IEEEtran}
\usepackage{amsmath,amsfonts}
\usepackage{array}
\usepackage{float}
\usepackage[utf8]{inputenc}
\usepackage[table]{xcolor}
\usepackage[linesnumbered,ruled]{algorithm2e}
\usepackage[left=0.625in, right=0.625in, top=0.75in, bottom=1in]{geometry}

\usepackage{booktabs}
\usepackage[english]{babel}

\usepackage{amsmath, amssymb, amsfonts, amsthm,algpseudocode}
\usepackage{diagbox} 
\usepackage{enumitem}
\usepackage{forest}
\usepackage{geometry, graphicx}
\usepackage{hyperref}
\usepackage{longtable}
\usepackage{multirow, makecell}
\usepackage{newunicodechar}
\usepackage{pgf}
\usepackage{subcaption}
\usepackage{caption}
\usepackage{tabularx, tikz}
\usepackage{verbatim}

\newunicodechar{​}{}
\usepackage{balance}

\definecolor{pink}{RGB}{244, 183, 190} 
\definecolor{deep_pink}{RGB}{239,148,159}
\definecolor{light_green}{RGB}{227,242,217} 
\definecolor{blue}{RGB}{208,228,246} 
\definecolor{deep_orange}{RGB}{255,136,16}
\definecolor{orange}{RGB}{244,179,130} 
\definecolor{blue_green}{RGB}{210,244,242} 
\definecolor{light_yellow}{RGB}{254,231,150} 
\definecolor{light_gray}{RGB}{166,166,166}
\usepackage{placeins}
\def\BibTeX{{\rm B\kern-.05em{\sc i\kern-.025em b}\kern-.08em
    T\kern-.1667em\lower.7ex\hbox{E}\kern-.125emX}}
    
\title{Text Data Augmentation for Large Language Models: A Comprehensive Survey of Methods, Challenges, and Opportunities}
\author{Yaping Chai, Haoran Xie, Joe S. Qin
\thanks{This work was supported by a grant from the Research Grants Council of the Hong Kong Special Administrative Region, China (R1015-23) and the Faculty Research Grant (SDS24A8) and the Direct Grant (DR25E8) of Lingnan University, Hong Kong. \emph{(Corresponding author: Haoran Xie.)}}
\thanks{Yaping Chai, Haoran Xie, and Joe S. Qin are with the School of Data Science, Lingnan University, Hong Kong (e-mail: yapingchai@ln.hk; hrxie@ln.edu.hk; joeqin@ln.edu.hk).}
}
\date{XXX}

\begin{document}
\maketitle

\begin{abstract}
The increasing size and complexity of pre-trained language models have demonstrated superior performance in many applications, but they usually require large training datasets to be adequately trained. Insufficient training sets could unexpectedly make the model overfit and fail to cope with complex tasks. Large language models (LLMs) trained on extensive corpora have prominent text generation capabilities, which improve the quality and quantity of data and play a crucial role in data augmentation. Specifically, distinctive prompt templates are given in personalised tasks to guide LLMs in generating the required content. Recent promising retrieval-based techniques further improve the expressive performance of LLMs in data augmentation by introducing external knowledge to enable them to produce more grounded-truth data. This survey provides an in-depth analysis of data augmentation in LLMs, classifying the techniques into Simple Augmentation, Prompt-based Augmentation, Retrieval-based Augmentation and Hybrid Augmentation. We summarise the post-processing approaches in data augmentation, which contributes significantly to refining the augmented data and enabling the model to filter out unfaithful content. Then, we provide the common tasks and evaluation metrics. Finally, we introduce existing challenges and future opportunities that could bring further improvement to data augmentation.
\end{abstract}


\begin{figure*}[htbp]
    \centering
    \includegraphics[width=0.90\textwidth]{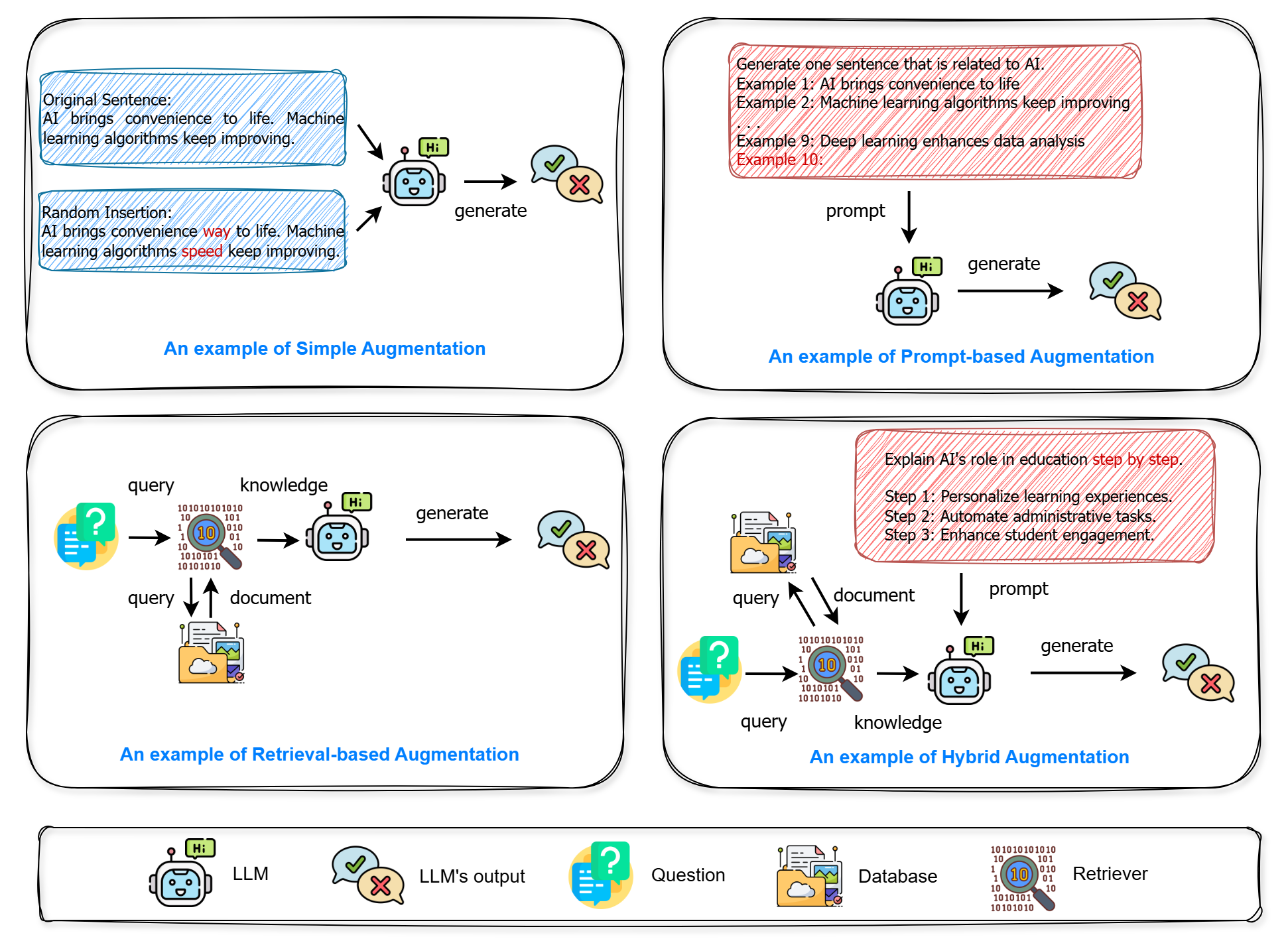}
    \caption{Four categories of data augmentation techniques.}
    \label{fig:DA}
\end{figure*}

\begin{figure*}[htbp]
    \centering
    \includegraphics[width=0.90\textwidth]{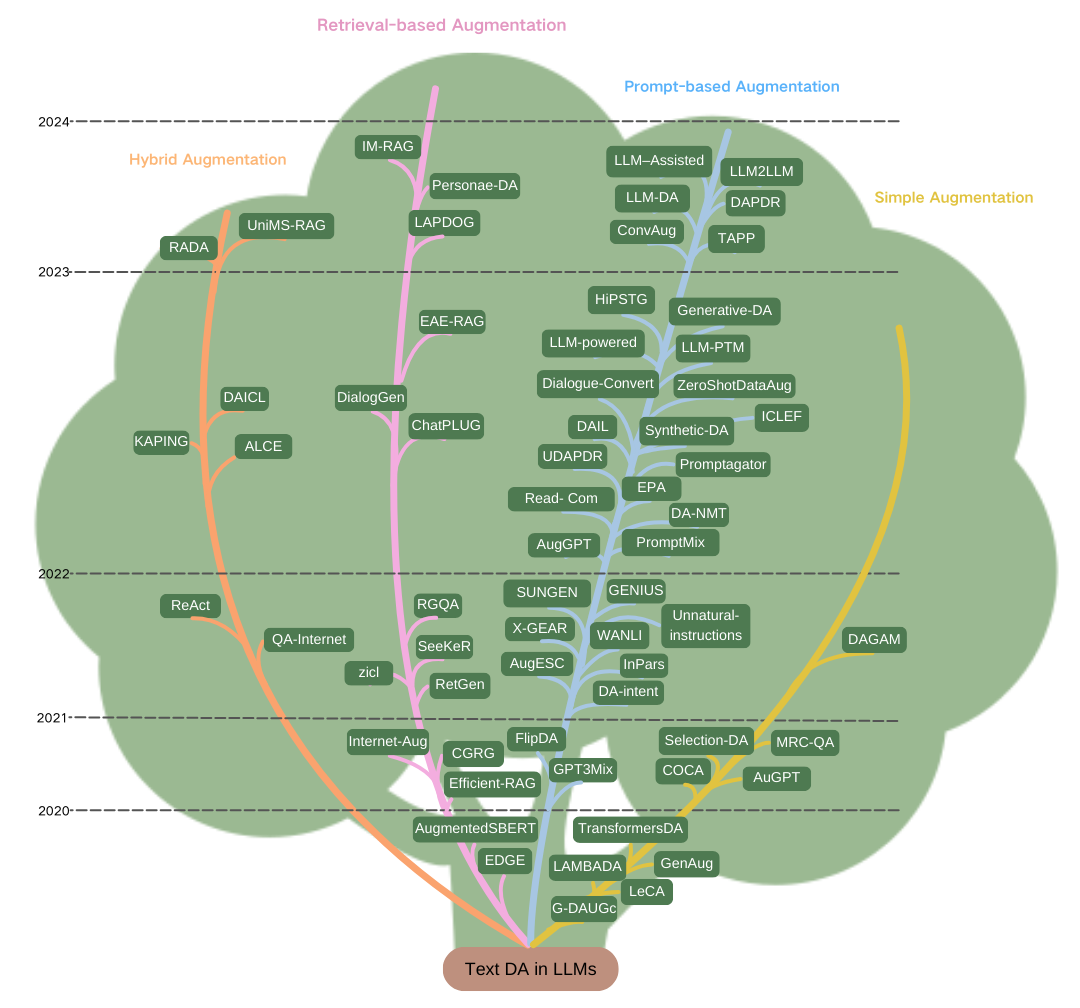}
    \caption{Recent studies on four categories of data augmentation techniques. As mentioned in section \ref{sec-DAmethods}, \textcolor{deep_orange}{Hybrid Augmentation} technique combines superior few-shot learning capabilities similar to prompt engineering and a retriever to obtain external knowledge. Using only the prompt portion of the RAG itself, we categorise it as the \textcolor{deep_pink}{Retrieval-based Augmentation} technique.}
    \label{fig:DAtree}
\end{figure*}

\section{Introduction}
\label{sec-Introduction}
\input{sec-Introduction}

\section{Preliminaries}
\label{sec-Preliminaries}

\input{Table/DAaspects}
\input{sec-Preliminaries}

\input{Table/DAlevel}

\input{sec-DAgranularity}

\section{Data augmentation techniques for text in LLMs}
\label{sec-DAmethods}

\begin{figure*}[htbp]
    \centering
    \includegraphics[width=0.95\textwidth]{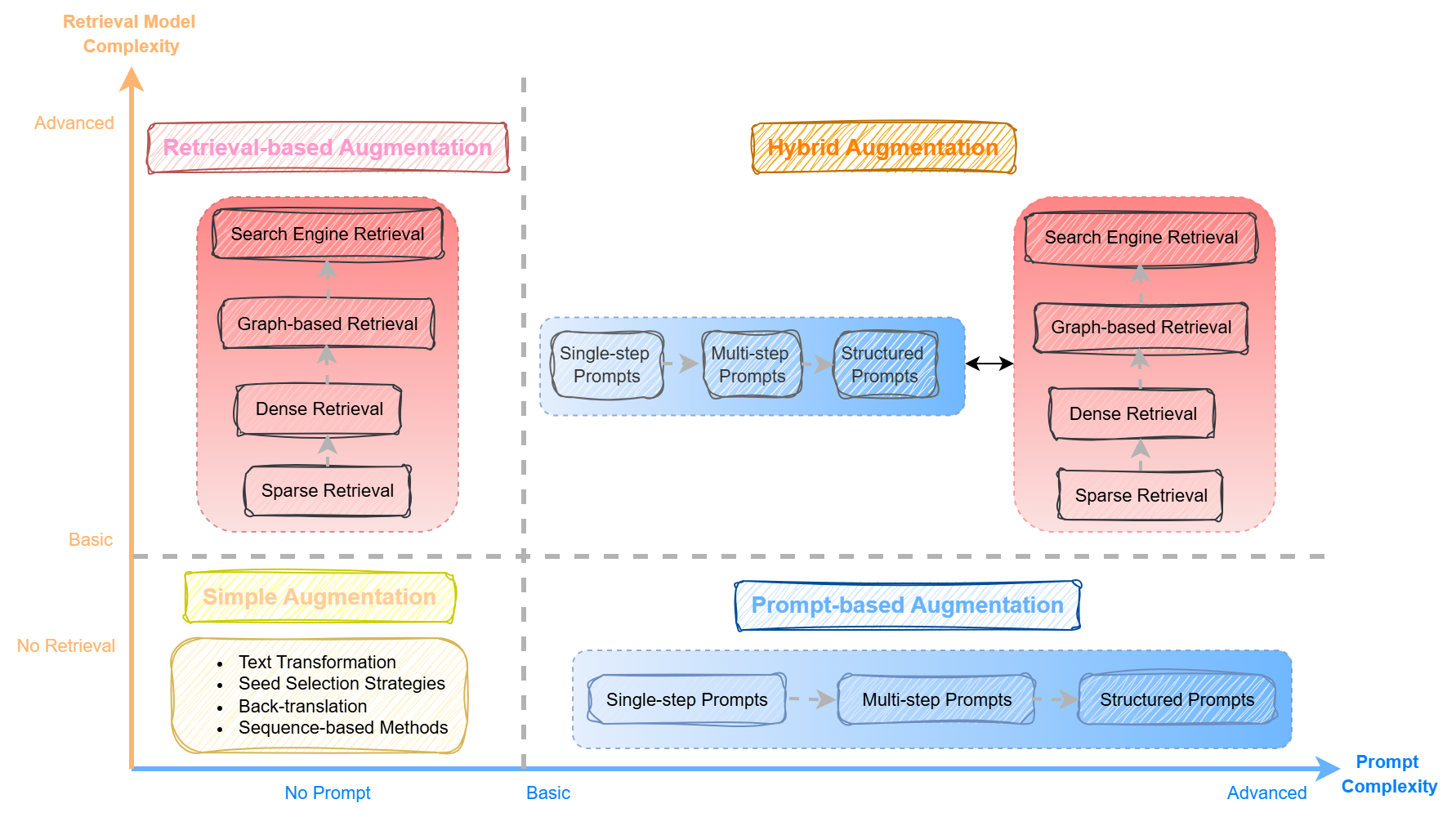}
    \caption{Data augmentation techniques in the  \textit{No Prompt-Basic-Advanced} spectrum according to Prompt Complexity and \textit{No Retrieval-Basic-Advanced} spectrum according to Retrieval Model Complexity.}
    \label{fig:spectrum}
\end{figure*}

\input{Figure/Fig-tree}

\input{sec-Pmethods}

\input{Table/Task}

\input{sec-Rmethods}

\section{Post-processing Approaches}
\label{sec-Post-processing}
\input{sec-Post-processing}
\input{Table/Evaluation}
\section{Tasks and Evaluation}
\label{sec-app}
\input{sec-applications}

\section{Challenges and opportunities}
\label{sec-Challenge}
\input{sec-Challenge}

\section{Conclusion}
\label{sec-CONCLUSION}
\input{sec-CONCLUSION}

\bibliographystyle{IEEEtran}
\bibliography{main}
\balance

\end{document}

%% file: sec-Introduction.tex
LLMs have demonstrated remarkable language understanding and generation capabilities with their massive number of parameters and training data, bringing innovation to many applications \cite{LLM-powered}. However, training and optimizing pre-trained language models (PLMs) requires enormous amounts of high-quality data. Poor data quality and data scarcity issues hinder the further improvement of PLMs \cite{AugGPT}. To solve these challenges, data augmentation is an effective technique to generate more available training data by transforming and expanding existing data, improving models' performance in many natural language processing (NLP) tasks \cite{LLM–Assisted}. The earliest popular data augmentation methods include synonym replacement, word order scrambling and random word deletion \cite{FlipDA}. Although these conventional methods could enhance existing data, simple word transformations may not be able to fully utilize the potential of LLMs \cite{GENIUS}. 

With the increasing development of prompt engineering and its application in LLMs, many researchers augment the seed training data by designing crafted prompts for LLMs to generate diverse datasets, contributing to producing high-quality and semantically rich data \cite{GPT-3}.

Providing prompts to LLMs cannot access up-to-date factual knowledge, unavoidably causing the hallucination phenomenon \cite{RetGen,QA-Internet}. Many studies have explored retrieving external data for data augmentation. Integrating additional knowledge significantly enhances PLMs' understanding and generalization capabilities by accessing relevant facts from a large-scale external corpus \cite{RAGsurvey}.

Along with the continuous progress of research, studies have not only retrieved grounded facts from external corpora but also combined them with LLMs' powerful few-shot capability to effectively apply them to various tasks \cite{QA-Internet}. A typical application is problem-answering tasks, where a retrieval model is used to obtain relevant documents from a database given a query. Then, adopting few-shot learning to prompt the model to answer questions based on the retrieved documents without fine-tuning or learning additional parameters \cite{QA-Internet}.

Based on the background of LLMs, we divide the techniques of text data augmentation into four categories: Simple Augmentation, Prompt-based Augmentation, Retrieval-based Augmentation, and Hybrid Augmentation. Figure \ref{fig:DA} demonstrates each example of four techniques. We present recent research on the four categories of text data augmentation in Figure \ref{fig:DAtree}. We focus on summarising how recent studies have performed data augmentation under four technique categories. We introduce a comprehensive review of multiple data augmentation aspects and granularity \cite{RAGsurvey}, aiming to illustrate the foundational preliminaries. Considering that the effectiveness of data augmentation cannot be entirely guaranteed, we demonstrate common post-processing approaches to refine generated data quality further. 

In summary, the contributions of this survey are as follows:
\begin{itemize}
\item We comprehensively demonstrate data augmentation techniques in the context of large-scale language models, including early Simple Augmentation, Prompt-based Augmentation, Retrieval-based Augmentation, and Hybrid Augmentation, outlining the development stages of data augmentation under the trend of increasing model size. 
\item We present two prerequisites before data augmentation, aspect and granularity, and introduce the refinement approaches after data augmentation, providing a systematic and complete structure.
\item We summarize the common tasks and evaluation metrics of data augmentation in various applications and analyze the current limitations and worthwhile future research directions.
\end{itemize}

The rest of this survey is structured as follows: section \ref{sec-Preliminaries} introduces LLMs, which typically refer to encoder-only models, decoder-only models, and encoder-decoder models. Followed by presenting data augmentation aspects and granularity from the finest to the coarsest. Section \ref{sec-DAmethods} showcases a detailed breakdown of the four technique categories, revealing typical traditional methods, the complexity of distinct designs for prompts, diverse retriever types and the various combinations of Prompt-based Augmentation and Retrieval-based Augmentation in recent studies. Section \ref{sec-Post-processing} focuses on interpreting widely used post-processing approaches for data augmentation. Section \ref{sec-app} describes common tasks and evaluation metrics. Section \ref{sec-Challenge} explains the unsolved challenges and future research directions. Finally, we conclude this survey in section \ref{sec-CONCLUSION}.

%% file: Table/DAaspects.tex
   		\begin{table*}[!ht] 		
			\caption{Data Augmentation Aspects.}
   \label{tab:Aspect}
    \centering
    \scalebox{1.0}{
				\begin{tabular}{c c c c c c c c} 
					\toprule
Methods                                                               & Generation & Paraphrasing & Translation & Labeling & Retrieval & Editing \\
\midrule
\multicolumn{7}{c}{Simple Augmentation}                                                                                             &  \\
\hline
TransformersDA \cite{TransformersDA}                 & \textbf{\checkmark}          &              &             &          &           &         \\
DAGAM \cite{DAGAM}                                   & \textbf{\checkmark}          &              &             &          &           & \textbf{\checkmark}       \\
GenAug \cite{GenAug}                                 &            &              &             &          &           & \textbf{\checkmark}       \\
AuGPT \cite{AuGPT}                                   &            &              & \textbf{\checkmark}           &          &           &         \\
COCA \cite{COCA}                                     &            &              &             &          &           & \textbf{\checkmark}       \\
Selection-DA \cite{Human-in-the-loop}           & \textbf{\checkmark}          &              &             &          &           &         \\
LAMBADA \cite{LAMBADA}                               & \textbf{\checkmark}          &              &             & \textbf{\checkmark}        &           &         \\
LeCA \cite{LeCA}                                     & \textbf{\checkmark}          &              &             &          &           &         \\
G-DAUGc \cite{G-DAUGc}                               & \textbf{\checkmark}          &              &             &          &           &         \\
MRC-QA  \cite{MRC-QA}                                &          &              &             &          &           & \textbf{\checkmark}       \\
\hline
\multicolumn{7}{c}{Prompt-based Augmentation}                                                                                                    \\
\hline
GPT3Mix \cite{GPT3Mix}                               & \textbf{\checkmark}          &              &             &          &           &         \\
DA-intent \cite{DA-intent}                           & \textbf{\checkmark}          &              &             &          &           &         \\
WANLI \cite{WANLI}                                   & \textbf{\checkmark}          &              &             &          &           &         \\
FlipDA \cite{FlipDA}                                 & \textbf{\checkmark}          &              &             & \textbf{\checkmark}        &           &         \\
AugESC \cite{AugESC}                                 & \textbf{\checkmark}          &              &             &          &           &         \\
AugGPT \cite{AugGPT}                                 &            & \textbf{\checkmark}            &             &          &           &         \\
Read-Com \cite{Read-Com}                            & \textbf{\checkmark}          &              &             &          &           &         \\
DAIL \cite{DAIL}                                     &            & \textbf{\checkmark}            &             & \textbf{\checkmark}        &           &         \\
DA-NMT \cite{DA-NMT}                                 & \textbf{\checkmark}          & \textbf{\checkmark}            & \textbf{\checkmark}           &          &           &         \\
EPA \cite{EPA}                                       &            & \textbf{\checkmark}            &             &          &           &         \\
ZeroShotDataAug \cite{ZeroShotDataAug}               & \textbf{\checkmark}          &              &             &          &           &         \\
Dialogue-Convert \cite{Dialogue-Convert}             & \textbf{\checkmark}          &              &             &          &           &         \\
HiPSTG \cite{HiPSTG}                                 & \textbf{\checkmark}          &              &             &          &           &         \\
SUNGEN \cite{PROMPTING}                              & \textbf{\checkmark}          &              &             &          &           &         \\
LLM-powered \cite{LLM-powered}                       & \textbf{\checkmark}          &              &             &          &           &         \\
LLM-PTM \cite{Healthcare-DA}                         & \textbf{\checkmark}          &              &             &          &           &         \\
Generative-DA \cite{Generative-DA}                   & \textbf{\checkmark}          &              &             &          &           &         \\
ICLEF \cite{ICLEF}                                   & \textbf{\checkmark}          & \textbf{\checkmark}            &             &          &           &         \\
LLM-DA \cite{LLM-DA}                                 & \textbf{\checkmark}          &              &             &          &           & \textbf{\checkmark}       \\
Synthetic-DA \cite{Synthetic-DA}                     & \textbf{\checkmark}          &              &             & \textbf{\checkmark}        &           &         \\
LLM–Assisted \cite{LLM–Assisted}                     & \textbf{\checkmark}          &              &             &          &           &         \\
LLM2LLM \cite{LLM2LLM}                               & \textbf{\checkmark}          &              &             &          &           &         \\
PromptMix \cite{PromptMix}                           & \textbf{\checkmark}          &              &             & \textbf{\checkmark}        &           &         \\
Unnatural-instructions \cite{Unnatural-instructions} & \textbf{\checkmark}          &              &             &          &           &         \\
GENIUS \cite{GENIUS}                                 & \textbf{\checkmark}          &              &             &          &           &         \\
TAPP \cite{Prefix-Prompt}                            & \textbf{\checkmark}          &              &             &          &           &         \\
X-GEAR \cite{X-GEAR}                                 & \textbf{\checkmark}          &              &             &          &           &         \\
InPars \cite{InPars}                                 & \textbf{\checkmark}          &              &             & \textbf{\checkmark}        &           &         \\
ConvAug \cite{ConvAug}                               &            & \textbf{\checkmark}            &             &          &           & \textbf{\checkmark}       \\
Promptagator \cite{Promptagator}                     & \textbf{\checkmark}          &              &             &          &           &         \\
DAPDR \cite{DAPDR}                                   & \textbf{\checkmark}          &              &             &          &           &         \\
UDAPDR \cite{UDAPDR}                                 & \textbf{\checkmark}          &              &             &          &           &         \\
\hline
\multicolumn{7}{c}{Retrieval-based Augmentation}                                                                                                 \\
\hline
AugmentedSBERT \cite{AugmentedSBERT}                 &            &              &             & \textbf{\checkmark}        & \textbf{\checkmark}         &         \\
zicl \cite{zicl}                                     &            &              &             & \textbf{\checkmark}        & \textbf{\checkmark}         &         \\
RetGen  \cite{RetGen}                                &            &              &             &          & \textbf{\checkmark}         &         \\
Internet-Aug  \cite{Internet-Aug}                    &            &              &             &          & \textbf{\checkmark}         &         \\
DialogGen  \cite{DialogGen}                          &            &              &             &          & \textbf{\checkmark}         &         \\
ChatPLUG \cite{ChatPLUG}                             &            &              &             &          & \textbf{\checkmark}         &         \\
EDGE  \cite{EDGE}                                    &            &              &             &          & \textbf{\checkmark}         &         \\
RGQA \cite{RGQA}                                     &            &              &             &          & \textbf{\checkmark}         &         \\
CGRG  \cite{CGRG}                                    &            &              &             &          & \textbf{\checkmark}         &         \\
IM-RAG \cite{IM-RAG}                                 &            &              &             &          & \textbf{\checkmark}         &         \\
EAE-RAG \cite{EAE-RAG}                               &            &              &             &          & \textbf{\checkmark}         &         \\
SeeKeR \cite{SeeKeR}                                 &            &              &             &          & \textbf{\checkmark}         &         \\
Efficient-RAG \cite{Efficient-RAG}                   &            &              &             &          & \textbf{\checkmark}         &         \\
LAPDOG \cite{LAPDOG}                                 &            &              &             &          & \textbf{\checkmark}         &         \\
Personae-DA \cite{Personae-DA}                       &            &              &             &          & \textbf{\checkmark}         &         \\
\hline
\multicolumn{7}{c}{Hybrid Augmentation}                                                                                                          \\
\hline
DAICL \cite{DAICL}                                   & \textbf{\checkmark}          &              &             &          & \textbf{\checkmark}         &         \\
KAPING \cite{KAPING}                                 &            &              &             &          & \textbf{\checkmark}         &         \\
ALCE \cite{ALCE}                                     &            &              &             &          & \textbf{\checkmark}         &         \\
RADA \cite{RADA}                                     & \textbf{\checkmark}          &              &             &          & \textbf{\checkmark}         &         \\
UniMS-RAG \cite{UniMS-RAG}                           &            &              &             &          & \textbf{\checkmark}         &         \\
QA-Internet \cite{QA-Internet}                       & \textbf{\checkmark}          &              &             &          & \textbf{\checkmark}         &         \\
ReAct \cite{ReAct}                                   &            &              &             &          & \textbf{\checkmark}         &     \\   
\bottomrule
				\end{tabular}
			}	
		
			\end{table*}

%% file: sec-Preliminaries.tex
\subsection{Large Language Models} 
	Large language models (LLMs) trained on an enormous corpus with rich language knowledge and prominent generation capabilities are widely used in academia and industry. A popular approach to divide LLMs is based on the Transformer \cite{Transformer} architecture: encoder-only models (e.g. the BERT families \cite{BERT}), decoder-only models (e.g. the GPT families \cite{GPT}), and encoder-decoder models (e.g. T5 \cite{T5} and BART \cite{BART}).
	\subsubsection{Encoder-only Models}
	The encoder-only model is represented by BERT \cite{BERT}, which has the main feature of bidirectional encoding. BERT performs contextual bidirectional encoding on the input data and captures multiple semantic information of each word in different contexts, enabling BERT and its variants to showcase outstanding performance in downstream tasks such as sentiment classification \cite{TransformersDA,GPT3Mix,ABSA} named entity recognition \cite{LLM-DA,GENIUS}, and question-answering systems \cite{Read-Com,Generative-DA,FlipDA}.
	
	\subsubsection{Decoder-only Models}
	Compared with encoder-only models, decoder-only models include ChatGPT \cite{GPT-3.5} that only use the Transformer decoder structure and demonstrate powerful text generation abilities. They predict text word by word through unidirectional decoding and produce coherent and natural responses, allowing them to generate diverse and high-quality training data that effectively improve models' robustness and generalisation ability \cite{LLM–Assisted,LLM2LLM,Prefix-Prompt}.
	
	\subsubsection{Encoder-decoder Models}
 Encoder-decoder models combine the advantages of Transformer's encoder and decoder. Representative models are T5 \cite{T5} and BART \cite{BART}. The dual-module structure empowers them not only to understand the deep semantics of the input content but also to generate semantically augmented data. Many researches train encoder-decoder models for diverse NLP tasks \cite{DA-NMT,Dialogue-Convert,ICLEF,Synthetic-DA,Unnatural-instructions} and utilize them for text generation without fine-tuning \cite{FlipDA,DAGAM,DAIL,Generative-DA}.

\subsection{Data Augmentation Aspects}
			Data augmentation (DA) is one of the critical techniques to improve the performance of pre-trained language models by transforming existing data to obtain diverse and augmented data. Instead of limiting one DA aspect, combining multiple aspects could generate an extensive synthetic dataset. The common aspects of data augmentation include the following ways and are illustrated in Table \ref{tab:Aspect}.
\subsubsection{Data Generation}
			Data generation is one of the most basic and crucial aspects of data augmentation \cite{DAsurvey}. A growing trend in generation tasks is to employ a Transformer-based language model with a decoder component to create new datasets \cite{DAGAM}. 
   
   For decoder-only models, such as Read-Com \cite{Read-Com} leverages GPT-4 \cite{GPT-4} to obtain synthesised datasets similar to the original data's style and semantics. LLM–Assisted \cite{LLM–Assisted} adopts Llama2 \cite{Llama2} to produce three different level augmentations based on a provided sample in the training set. 
   
   For encoder-decoder models, DAGAM \cite{DAGAM} summarises a vast set of sentences from original datasets and utilizes T5 \cite{T5} to generate synthetic data with a similar representation distribution to original datasets. TransformersDA \cite{TransformersDA} first employs two word-level masking strategies to reconstruct the original sequence. Then, it applies BART \cite{BART} to decode the masked sequence and produce high-quality output text.

			\subsubsection{Data Paraphrasing}
			Data paraphrasing refers to expressing the original training data differently within a reasonable deviation. The original training data is rephrased by changing the sentence's structure, wording or grammar. EPA \cite{EPA} proposes to paraphrase the available demonstrations on the source and target sides to acquire generated demonstrations for in-context learning. DAIL \cite{DAIL} expands the data by using the LLM itself to paraphrase the original test samples and obtain multiple candidate texts. The final labels are created by not only considering the original text but also taking into account all paraphrased candidates. Unnatural-instructions \cite{Unnatural-instructions} first use structured instruction format and some filtering heuristic methods to collect a core sample set. Then, it rephrases structured instructions to extend the core dataset by prompting the language model to use manually constructed samples. Similarly, AugGPT \cite{AugGPT} designs prompts to guide LLMs to rephrase the original data and obtain more data. ConvAug \cite{ConvAug} introduces to mimic the variety of user expressions for similar intentions by utilising the LLM to extend the language's diversity by paraphrasing the whole context, reducing the model to overfit specific phrases or patterns.

			\subsubsection{Data Translation}
			Data translation aims to translate text from one language to another language and then translate it back to the original language. DA-NMT \cite{DA-NMT} generates $n$ translations with the same meaning but different languages for each original source sentence $s$, then maps a source sentence to $n$ target sentences, and finally generates $n$ parallel data. AuGPT \cite{AuGPT} utilizes the advantages of back-translation \cite{translation} and employs a well-trained multilingual machine translation model to rephrase the data. It uses ten intermediate languages to obtain distinct new data for each input utterance.
			
			\subsubsection{Data Editing}
			Data editing refers to performing simple transformations on the text, which includes insertion\cite{GenAug}, reordering \cite{COCA}, deletion \cite{GenAug}, rewriting \cite{LLM-DA} and shuffling  \cite{DAGAM}. DAGAM \cite{DAGAM} implements a \textit{character order change} strategy by fixing the first and last characters in a word and shuffling the rest to augment the data. LLM-DA \cite{LLM-DA} creates entities that are semantically similar to the sentence but diverse by replacing entities in the sentence with other entities of the same type. ConvAug \cite{ConvAug} proposes Entity Replacing as one of its augmentation strategies. It uses the LLM to recognise and replace entities in a context similar to the original but differing in significant details. The model could focus more closely on crucial information rather than trivial aspects through contrastive learning. Many tasks assume a strict sequence order. However, COCA \cite{COCA} observes that the sequence of users' search behaviour is more flexible. It proposes a behavioural reordering strategy to avoid relying too much on sequential information.
   
			\subsubsection{Data labeling}
			Data labeling refers to assigning labels to synthetic samples or unlabeled data \cite{DAsurvey}. In practice, adopting manual annotation to obtain more ground-truth labels is expensive and cannot be performed on a considerable scale \cite{Synthetic-DA}. Manual annotation by experts may require additional implicit knowledge that is difficult to capture through statistical methods \cite{Human-in-the-loop}.

   A common way to automatically label synthesised examples is using a language model \cite{DAIL}. In the domain adaptation data augmentation strategy, AugmentedSBERT \cite{AugmentedSBERT} first fine-tune BERT on the source domain comprising paired annotations. After fine-tuning, the fine-tuned BERT is applied to annotate the target domain. Finally, training SBERT on the annotated target domain sentence pairs. Once the retrieved sentences are obtained, zicl \cite{zicl} pairs each sentence with a label by a synonym labeling strategy to reduce the \textit{copy effect} phenomenon. DAIL \cite{DAIL} considers candidate labels from the original sample and all paraphrased texts and determines the final label by majority voting method. Synthetic-DA \cite{Synthetic-DA} attempts two scenarios: one is to annotate the set of unlabeled instances when there is a significant amount of unlabeled text, but the cost of obtaining annotations is high; the other is to generate the entire input-output pair without available annotated instances. PromptMix \cite{PromptMix} first directs an LLM to produce samples that mix two classes and then uses relabeling prompts to enhance the faithfulness of the synthesised examples.
			
   \subsubsection{Data Retrieval}
			Data retrieval is an efficient way to use external data to retrieve samples related to the current task from a large-scale text corpus and concatenate them into the training set \cite{RAGsurvey}. AugmentedSBERT \cite{AugmentedSBERT} and zicl \cite{zicl} expand the data volume by incorporating data retrieved from the external source with the original data. DAICL \cite{DAICL} combines each input from the source domain with relevant retrieved texts from the target domain to enrich the dataset. Then, the model learns task discrimination based on the source content and target texts. 
   
   Rather than incorporating the retrieved content with existing data for training, more studies have incorporated relevant information and ground truth retrieved from the external corpus into LLM's input context, thereby enhancing LLM's factual response \cite{RADA}. KAPING \cite{KAPING} and ALCE \cite{ALCE} employ the retrieved information as additional knowledge to guide the model to produce more accurate responses. 
   
   Also, recent researches explore utilising search engines \cite{DialogGen} and various APIs \cite{ReAct,QA-Internet,SeeKeR} to conduct retrieval and access up-to-date knowledge from a more extensive source.

%% file: Table/DAlevel.tex
			\begin{table*}[!ht]
				\caption{Data Augmentation Granularity.}
    	\label{tap:DAlevel}
     \centering
        \scalebox{1.0}{
\begin{tabular}{ccccccc}
\toprule
Methods                                                               & Token Level & Token-span Level & Sentence Level & Passage Level & Context Level & Document Level \\
\midrule
\multicolumn{7}{c}{Simple Augmentation}                                                                                                                                  \\
\hline
TransformersDA \cite{TransformersDA}                 & \textbf{\checkmark}           & \textbf{\checkmark}                &                &               &               &                \\
DAGAM \cite{DAGAM}                                   & \textbf{\checkmark}           &                  &                &               &               & \textbf{\checkmark}              \\
GenAug \cite{GenAug}                                 & \textbf{\checkmark}           &                  &                &               &               &                \\
AuGPT \cite{AuGPT}                                   &             &                  & \textbf{\checkmark}              &               &               &                \\
COCA \cite{COCA}                                     & \textbf{\checkmark}           &                  &                &               &               & \textbf{\checkmark}              \\
Selection-DA \cite{Human-in-the-loop}                &             &                  & \textbf{\checkmark}              &           &               &                \\
LAMBADA \cite{LAMBADA}                               & \textbf{\checkmark}           &                  & \textbf{\checkmark}              &               &               &                \\
LeCA \cite{LeCA}                                     & \textbf{\checkmark}           &                  & \textbf{\checkmark}              &               &               &                \\
G-DAUGc \cite{G-DAUGc}                               &             &                  & \textbf{\checkmark}              &               &               &                \\
MRC-QA  \cite{MRC-QA}                                &             & \textbf{\checkmark}                &                &               &               &                \\
\hline
\multicolumn{7}{c}{Prompt-based Augmentation}                                                                                                                            \\
\hline
 GPT3Mix \cite{GPT3Mix}                               &             &                  & \textbf{\checkmark}              &               &               &                \\
DA-intent \cite{DA-intent}                           &             &                  & \textbf{\checkmark}              &               &               &                \\
WANLI \cite{WANLI}                                   &             &                  & \textbf{\checkmark}              &               &               &                \\
FlipDA \cite{FlipDA}                                 & \textbf{\checkmark}           &                  & \textbf{\checkmark}              &               &               &                \\
AugESC \cite{AugESC}                                 &             &                  &                &               &               & \textbf{\checkmark}              \\
AugGPT \cite{AugGPT}                                 &             &                  & \textbf{\checkmark}              &               &               &                \\
Read- Com \cite{Read-Com}                           &             &                  &                &               & \textbf{\checkmark}             &                \\
DAIL \cite{DAIL}                                     &             &                  & \textbf{\checkmark}              &               &               &                \\
DA-NMT \cite{DA-NMT}                                 &             &                  & \textbf{\checkmark}              &               &               &                \\
EPA \cite{EPA}                                       &             &                  & \textbf{\checkmark}              &               &               &                \\
ZeroShotDataAug \cite{ZeroShotDataAug}               &             &                  & \textbf{\checkmark}              &               &               &                \\
Dialogue-Convert \cite{Dialogue-Convert}             &             &                  &                &               & \textbf{\checkmark}             &                \\
HiPSTG \cite{HiPSTG}                                 &             &                  &                &               &               & \textbf{\checkmark}              \\
SUNGEN \cite{PROMPTING}                              &             &                  & \textbf{\checkmark}              &               &               &                \\
LLM-powered \cite{LLM-powered}                       &             &                  &                &               & \textbf{\checkmark}             &                \\
LLM-PTM \cite{Healthcare-DA}                         &             &                  & \textbf{\checkmark}              &               &               &                \\
Generative-DA \cite{Generative-DA}                   &             &                  &                &               & \textbf{\checkmark}             &                \\
ICLEF \cite{ICLEF}                                   &             &                  & \textbf{\checkmark}              &               &               &                \\
LLM-DA \cite{LLM-DA}                                 &             & \textbf{\checkmark}                & \textbf{\checkmark}              &               &               &                \\
Synthetic-DA \cite{Synthetic-DA}                     &             &                  &                &               & \textbf{\checkmark}             &                \\
LLM–Assisted \cite{LLM–Assisted}                     & \textbf{\checkmark}           &                  &                &               &               & \textbf{\checkmark}              \\
LLM2LLM \cite{LLM2LLM}                               &             &                  &                &               & \textbf{\checkmark}             &                \\
PromptMix \cite{PromptMix}                           &             &                  & \textbf{\checkmark}              &               &               &                \\
Unnatural-instructions \cite{Unnatural-instructions} &             &                  &                &               & \textbf{\checkmark}             &                \\
GENIUS \cite{GENIUS}                                 &             & \textbf{\checkmark}                & \textbf{\checkmark}              &               &               &                \\
TAPP \cite{Prefix-Prompt}                            &             &                  &                &               & \textbf{\checkmark}             &                \\
X-GEAR \cite{X-GEAR}                                 & \textbf{\checkmark}           &                  &                & \textbf{\checkmark}             &               &                \\
InPars \cite{InPars}                                 &             &                  & \textbf{\checkmark}              &               &               &                \\
ConvAug \cite{ConvAug}                               & \textbf{\checkmark}           &                  &                &               & \textbf{\checkmark}             &                \\
Promptagator \cite{Promptagator}                     &             &                  & \textbf{\checkmark}              &               &               &                \\
DAPDR \cite{DAPDR}                                   &             &                  & \textbf{\checkmark}              &               &               &                \\
UDAPDR \cite{UDAPDR}                                 &             &                  & \textbf{\checkmark}              &               &               &                \\
\hline
\multicolumn{7}{c}{Retrieval-based Augmentation}                                                                                                                         \\
\hline
AugmentedSBERT \cite{AugmentedSBERT}                 &             &                  & \textbf{\checkmark}              &               &               &                \\
zicl \cite{zicl}                                     &             &                  & \textbf{\checkmark}              &               &               &                \\
RetGen  \cite{RetGen}                                &             &                  &                &               &               & \textbf{\checkmark}              \\
Internet-Aug  \cite{Internet-Aug}                    &             &                  &                &               &               & \textbf{\checkmark}              \\
DialogGen  \cite{DialogGen}                          &             &                  &                &               &               & \textbf{\checkmark}              \\
ChatPLUG \cite{ChatPLUG}                             &             &                  &                &               &               & \textbf{\checkmark}              \\
EDGE  \cite{EDGE}                                    &             &                  &                &               & \textbf{\checkmark}             &                \\
RGQA \cite{RGQA}                                     &             &                  &                &               & \textbf{\checkmark}             &                \\
CGRG  \cite{CGRG}                                    &             &                  & \textbf{\checkmark}              &               &               &                \\
IM-RAG \cite{IM-RAG}                                 &             &                  &                &               &               & \textbf{\checkmark}              \\
EAE-RAG \cite{EAE-RAG}                               &             &                  &                &               &               & \textbf{\checkmark}              \\
SeeKeR \cite{SeeKeR}                                 &             &                  &                &               &               & \textbf{\checkmark}              \\
Efficient-RAG \cite{Efficient-RAG}                   &             &                  &                &               &               & \textbf{\checkmark}              \\
LAPDOG \cite{LAPDOG}                                 &             &                  &                &               &               & \textbf{\checkmark}              \\
Personae-DA \cite{Personae-DA}                       &             &                  &                &               &               & \textbf{\checkmark}              \\
\hline
\multicolumn{7}{c}{Hybrid Augmentation}                                                                                                                                  \\
\hline
DAICL \cite{DAICL}                                   &             &                  &                &               & \textbf{\checkmark}             &                \\
KAPING \cite{KAPING}                                 &             &                  &                &               & \textbf{\checkmark}             &                \\
ALCE \cite{ALCE}                                     &             &                  &                & \textbf{\checkmark}             &               &                \\
RADA \cite{RADA}                                     &             &                  &                &               & \textbf{\checkmark}             &                \\
UniMS-RAG \cite{UniMS-RAG}                           &             &                  &                &               & \textbf{\checkmark}             &                \\
QA-Internet \cite{QA-Internet}                       &             &                  &                &               & \textbf{\checkmark}             &                \\
ReAct \cite{ReAct}                                   &             &                  & \textbf{\checkmark}              &               &               &  \\   
\bottomrule

\end{tabular}
}

\end{table*}

%% file: sec-DAgranularity.tex
\subsection {Data Augmentation Granularity}
			In addition to data augmentation aspects, data augmentation granularity significantly affects the augmented content. A coarser granularity typically less perturbs the original data and retains more initial information, while a finer granularity is more likely to produce more significant perturbations \cite{RAGsurvey}. Different data augmentation granularity could control the created data's diversity and preserve the original data's characteristics. Detailed information is depicted in Table \ref{tap:DAlevel} to present diverse granularity.
			
			\subsubsection{Token Level}
			Token level data augmentation is the most fine-grained granularity, mainly performed on single words. FlipDA \cite{FlipDA} first combines text $x$ and labels $y$ into a sequence using a cloze pattern \cite{cloze} and then randomly masks a fixed percentage of the input tokens. DAGAM \cite{DAGAM} proposes a \textit{character order change (COC)} strategy to achieve token insertion, token deletion and token replacement effects. RGQA \cite{RGQA} introduces three new tokens to demonstrate each component of the input sequence in the prompt and instantiate all portions to construct the final input sequence. ConvAug \cite{ConvAug} treats a context as a sequence of tokens \textit{C} and introduces two token level approaches on \textit{C}, which are Token Masking and Entity Replacing, to help the model learn subtle information differences. Inspired by the \textit{word mask} technique, COCA \cite{COCA} proposes to mask terms randomly over user behavioural sequences to prevent model overfitting and enhance the robustness of behaviour sentence representations.
			
			\subsubsection{Token-span Level}
			Compared with single token level augmentation, the token-span level involves a set of continuous tokens that could obtain more semantic information between continuous tokens \cite{SpanBERT}. TransformersDA \cite{TransformersDA} applies two masking strategies to fine-tune BART: one is replacing a single word with a mask token \textit{\texttt{<mask>}}, and the other is masking a continuous chunk of words with a single \textit{\texttt{<mask>}}. Motivated by the outstanding performance of predicting masked words \cite{SpanBERT} and contiguous spans \cite{T5} as self-supervised training objectives, Dialogue-Convert \cite{Dialogue-Convert} concatenates gap text spans into a pseudo-summary in the pre-training stage for the generation task in the medical field. KAPING \cite{KAPING} proposes to retrieve triples that are question-only related. One triple comprises $(s, r, o)$, where $s$ and $o$ represent the subject and object entities, respectively, and $r$ is a specific relation type between them. LLM-DA \cite{LLM-DA} adopts entity-level augmentation to tackle the NER task, which is necessary to understand the syntactic structure and semantic information of each token in the sentence to identify entities precisely. MRC-QA \cite{MRC-QA} firstly conducts probability sampling on which answers need to be employed for data augmentation. Secondly, once the question-answer pairs are selected, a set of fuzzy answer spans are generated by shifting the correct answer span by certain characters to the left or right. 
			
			\subsubsection{Sentence Level}
			Sentence level is the most widely used granularity in data augmentation, mainly focusing on the whole sentence, such as sentence generation and translation. Given an input sentence and its related entities, LLM-DA \cite{LLM-DA} aims to generate a collection of sentence variants. These variants include different aspects such as vocabulary usage, clauses, and expression styles. New sentences with rewritten context are obtained throughout the process while keeping the entities from the initial sentences. ZeroShotDataAug \cite{ZeroShotDataAug} leverages the LLM to produce a significant number of sentences for each task in a zero-shot setting, improving the model's performance in the low-resourced scenario. LAMBADA \cite{LAMBADA} fine-tunes the LLM to learn the training data patterns and produce synthesised sentences based on specific labels. Some studies paraphrase the original sentence to increase the training dataset's diversity \cite{AugGPT,DAIL,DA-NMT,EPA}. 
   
   In addition to common NLP tasks, in information retrieval (IR), InPars \cite{InPars} and Promptagator \cite{Promptagator} adopt query generation for data augmentation and improve retrieval accuracy by generating a query and retrieving documents related to the query in the document set. UDAPDR \cite{UDAPDR} and DAPDR \cite{DAPDR} prompt an LLM to generate an enormous number of synthesis queries for passages, improving retrieval accuracy. 
   
   An innovative approach is to employ external datasets to achieve data augmentation. zicl \cite{ zicl} and AugmentedSBERT \cite{AugmentedSBERT} augment the limited dataset by retrieving sentences from an external corpus related to the training sentences. Then, concatenate the retrieved and original sentences to expand the dataset. 
			
			\subsubsection{Passage Level}
			Passage level is a coarser level of data augmentation granularity. It could be adjusting the order of passages or inserting new content between passages. Given a question, ALCE \cite{ALCE} creates text while citing corresponding supporting passages from a large-scale retrieval corpus. For the event argument extraction task, X-GEAR \cite{X-GEAR} extracts information from the input passage to fill in the language-agnostic template and guides the model to create a target string based on the language-agnostic template.

			\subsubsection{Context Level}
			Context level data augmentation requires language models to generate responses that demand an understanding of the input context. The most common applications are question-answering and conversation-generation tasks. Synthetic-DA \cite{Synthetic-DA} create more training data by providing LLM with prompts to understand the text context. LLM2LLM \cite{LLM2LLM} devises different system prompts for each task, which enable users to utilize domain-specific knowledge in the dataset generation process, and it leverages in-context learning to create more relevant examples. Given a context, RADA \cite{RADA} aims to generate an extracted question-answer pair and requires obtaining the answer to the question from the provided context. Unnatural-instructions \cite{Unnatural-instructions} constructs the meta-prompt containing the in-context example's number and each example's constant. This method prompts the model to create an example based on the in-context demonstrations.
			
			\subsubsection{Document Level}
			Document level data augmentation is the coarsest-grained granularity, usually utilized for document generation and retrieval tasks. AugESC \cite{AugESC} introduces the implementation of the LLM to accomplish the entire dialogue for diverse topics. HiPSTG \cite{HiPSTG} proposes a hierarchical discipline structure to generate the whole proposal by comprehending the structured prompt. COCA \cite{COCA} proposes three augmentation strategies, one of which is document level. This strategy utilizes contrastive learning for query/document deletion to enhance the learning of user behaviour sequence representations. 
            
            As for document retrieval tasks, SeeKeR \cite{SeeKeR} employs the search engine to retrieve relevant documents and then keep only the top 5 documents to generate the knowledge response. RetGen \cite{RetGen} introduces a framework with a dense document retriever component to obtain the most significant documents and a knowledge-grounded text generator component to produce the target output.

%% file: Figure/Fig-tree.tex
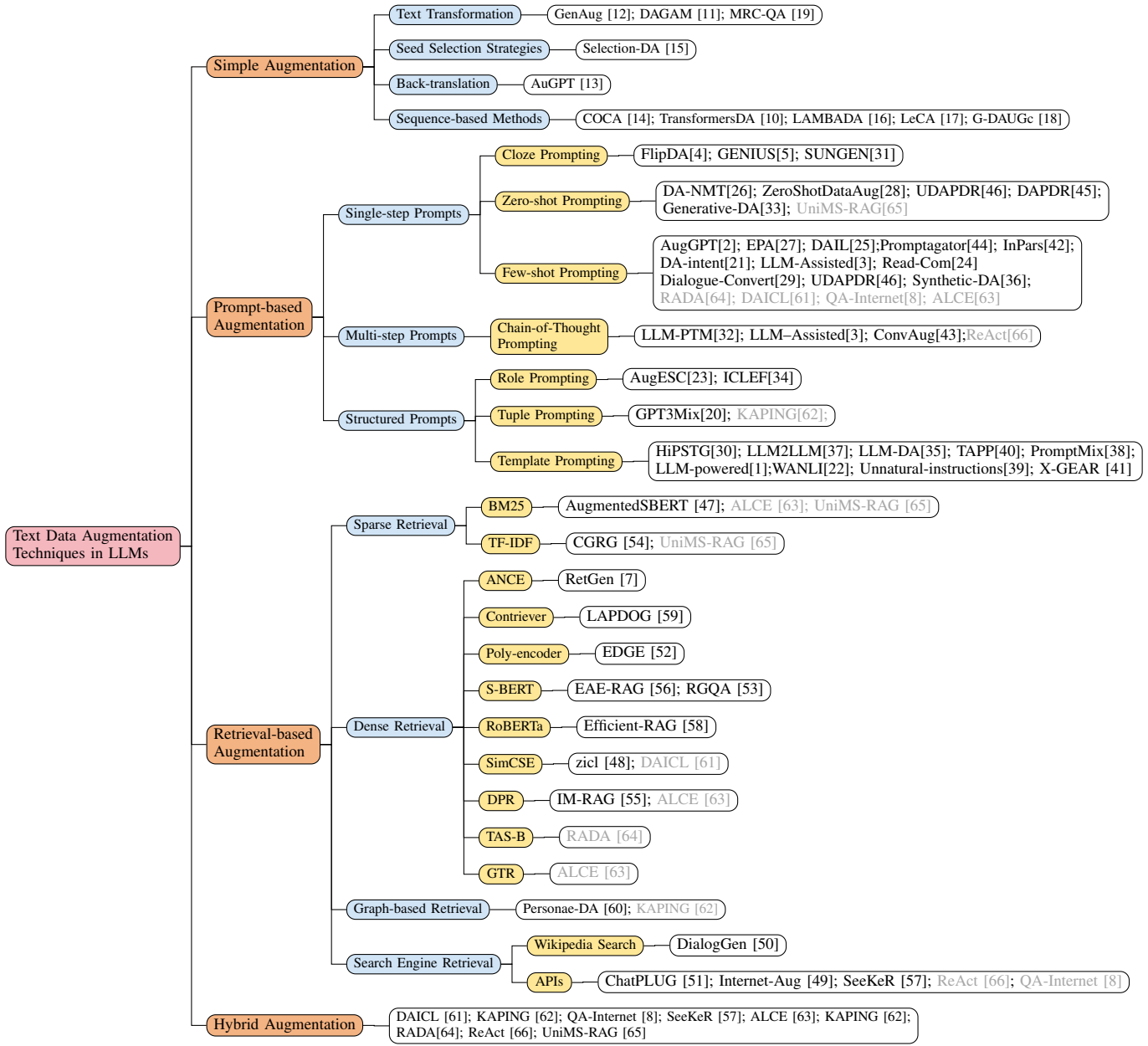
\begin{figure*}[htbp]
    \centering
\begin{forest}
  for tree={
  scale=0.8,
  align=center,
  grow=east,
  reversed=true,
  anchor=base west,
  parent anchor=east,
  child anchor=west,
  base=left,
  edge path={ 
      \noexpand\path [draw, \forestoption{edge}] (!u.parent anchor) -- +(5pt,0) |- (.child anchor)\forestoption{edge label};
    },
  font=\small,
  rectangle,
  draw,
  rounded corners,align=left,
  minimum width=2.5em,
  inner xsep=4pt,
  inner ysep=1pt,
  },
  where level=1{fill=orange}{},  
  where level=2{fill=blue, font=\footnotesize}{},  
  where level=3{fill=light_yellow, font=\footnotesize}{},  
[Text Data Augmentation \\Techniques in LLMs, fill=pink, 
        [Simple Augmentation
            [Text Transformation
            [GenAug \cite{GenAug}; DAGAM \cite{DAGAM}; MRC-QA \cite{MRC-QA}, fill=white]
            ] 
            [Seed Selection Strategies
            [Selection-DA \cite{Human-in-the-loop}, fill=white]
            ]          
            [Back-translation
            [AuGPT \cite{AuGPT}, fill=white]
            ]
            [Sequence-based Methods
            [COCA \cite{COCA}; TransformersDA \cite{TransformersDA}; LAMBADA \cite{LAMBADA}; LeCA \cite{LeCA}; G-DAUGc \cite{G-DAUGc}, fill=white]
            ]
        ]
        [Prompt-based\\Augmentation
            [Single-step Prompts
                [Cloze Prompting
                    [FlipDA\cite{FlipDA}; GENIUS\cite{GENIUS}; SUNGEN\cite{PROMPTING}]
               ]
                [Zero-shot Prompting
                    [DA-NMT\cite{DA-NMT}; ZeroShotDataAug\cite{ZeroShotDataAug}; UDAPDR\cite{UDAPDR}; DAPDR\cite{DAPDR};\\ Generative-DA\cite{Generative-DA}; \textcolor{light_gray}{UniMS-RAG\cite{UniMS-RAG}}]
               ]
               [Few-shot Prompting
                   [AugGPT\cite{AugGPT}; EPA\cite{EPA}; DAIL\cite{DAIL};Promptagator\cite{Promptagator};  InPars\cite{InPars}; \\ DA-intent\cite{DA-intent}; LLM-Assisted\cite{LLM–Assisted}; Read-Com\cite{Read-Com}\\
                   Dialogue-Convert\cite{Dialogue-Convert}; UDAPDR\cite{UDAPDR}; Synthetic-DA\cite{Synthetic-DA};\\\textcolor{light_gray}{RADA\cite{RADA};} \textcolor{light_gray}{DAICL\cite{DAICL};} \textcolor{light_gray}{QA-Internet\cite{QA-Internet};} \textcolor{light_gray}{ALCE\cite{ALCE}}]
               ]
            ]        
            [Multi-step Prompts
                [Chain-of-Thought\\ Prompting
                        [LLM-PTM\cite{Healthcare-DA}; LLM–Assisted\cite{LLM–Assisted}; ConvAug\cite{ConvAug};\textcolor{light_gray}{ReAct\cite{ReAct}}]                             
               ]
            ]
            [Structured Prompts
                [Role Prompting
                   [AugESC\cite{AugESC}; ICLEF\cite{ICLEF}]
                ]
                [Tuple Prompting
                    [GPT3Mix\cite{GPT3Mix}; \textcolor{light_gray}{KAPING\cite{KAPING};}]
                ]
                [Template Prompting
                    [HiPSTG\cite{HiPSTG}; LLM2LLM\cite{LLM2LLM}; LLM-DA\cite{LLM-DA}; TAPP\cite{Prefix-Prompt}; PromptMix\cite{PromptMix};\\ LLM-powered\cite{LLM-powered};WANLI\cite{WANLI}; Unnatural-instructions\cite{Unnatural-instructions}; X-GEAR \cite{X-GEAR}]
                ]
            ]
    ]
        [Retrieval-based\\Augmentation
            [Sparse Retrieval
            [BM25
                [AugmentedSBERT \cite{AugmentedSBERT};  \textcolor{light_gray}{ALCE \cite{ALCE};} \textcolor{light_gray}{UniMS-RAG  \cite{UniMS-RAG}}]
            ]
            [TF-IDF
                [CGRG \cite{CGRG}; \textcolor{light_gray}{UniMS-RAG  \cite{UniMS-RAG}}]
            ]
            ]
            [Dense Retrieval
                [ANCE, [RetGen  \cite{RetGen}]]
                [Contriever, [LAPDOG \cite{LAPDOG}]]
                [Poly-encoder, [EDGE \cite{EDGE}]]
                [S-BERT, [EAE-RAG \cite{EAE-RAG}; RGQA \cite{RGQA}]]
                [RoBERTa, [Efficient-RAG \cite{Efficient-RAG}]] 
                [SimCSE, [zicl \cite{zicl}; \textcolor{light_gray}{DAICL \cite{DAICL}}]]
                [DPR, [IM-RAG \cite{IM-RAG}; \textcolor{light_gray}{ALCE \cite{ALCE}}]]
                [TAS-B, [RADA \cite{RADA},text=light_gray]]
                [GTR, [ALCE \cite{ALCE},text=light_gray]]                               
            ] 
            [Graph-based Retrieval
                [Personae-DA \cite{Personae-DA}; \textcolor{light_gray}{KAPING \cite{KAPING}},fill=white]
            ]
            [Search Engine Retrieval
                [Wikipedia Search
                    [DialogGen  \cite{DialogGen}]
                ]
                [APIs
                    [ChatPLUG  \cite{ChatPLUG}; Internet-Aug  \cite{Internet-Aug}; SeeKeR \cite{SeeKeR}; \textcolor{light_gray}{ReAct \cite{ReAct}}; \textcolor{light_gray}{QA-Internet \cite{QA-Internet}}]
                ]
            ]              
        ]
    [Hybrid Augmentation [DAICL \cite{DAICL}; KAPING \cite{KAPING}; QA-Internet \cite{QA-Internet}; SeeKeR \cite{SeeKeR}; ALCE \cite{ALCE}; KAPING \cite{KAPING}; \\ RADA\cite{RADA}; ReAct \cite{ReAct}; UniMS-RAG \cite{UniMS-RAG}, fill=white]]
    ]
\end{forest}

\caption{Detailed data augmentation methods for four techniques. For a better understanding, the \textcolor{light_gray}{grey font} means that the paper is from \textcolor{light_gray}{Hybrid Augmentation}, and we could see from the figure how \textcolor{light_gray}{Hybrid Augmentation} designs the prompt and performs the retrieval.}
\label{fig:DA_methods}
\end{figure*}

%% file: sec-Pmethods.tex
Early traditional data augmentation methods include basic text transformation \cite{GenAug, DAGAM, MRC-QA}, and Back-translation \cite{AuGPT}. The arrival of large language models (LLMs) has given rise to an increasing trend of utilizing the power of generative models and prompts them to generate diverse datasets \cite{PE}. To address LLMs' inevitable limitations, such as hallucinations, retrieving relevant facts from extensive external corpora effectively improves the generated response \cite{RAG}. With the continuous progress of research, combining up-to-date knowledge with LLM's prominent few-shot learning capability has jointly promoted the application of data augmentation in various tasks \cite{RADA, DAICL, ALCE}. We classify data augmentation techniques into four categories based on the critical technologies used in recent years. Each technique has distinct characteristics, as Figure \ref{fig:spectrum} illustrates. We classify four DA techniques based on Prompt Complexity and Retrieval Model Complexity. 

\textbf{Simple Augmentation} is a conservative augmentation method that does not construct prompts or utilize retrieval methods. The augmented data is only slightly modified based on the original data and has a high similarity with the existing data \cite{GENIUS}. Similar to Simple Augmentation, \textbf{Prompt-based Augmentation} does not employ a retrieval module, but the difference is that it is a widely used technique during the flourishing period of prompt engineering, which opened up guiding large-scale language models to generate an extensive set of data by designing diverse prompts \cite{GPT3Mix}. Based on the Prompt Complexity, we categorize the recent prompt-based studies into basic Single-step Prompting (such as Zero-shot Prompting), Multi-step Prompting (such as Chain-of-Thought Prompting), and advanced Structured Prompting. In contrast, \textbf{Retrieval-based Augmentation} does not utilize LLM’s few-shot learning ability, but instead obtains external knowledge through a retrieval model. The Retrieval Model Complexity ranges from Sparse Retrieval and Dense Retrieval in the early information retrieval stage \cite{IFsurvey}, to more complex Graph-based Retrieval, and now with the emergence of RAG \cite{RAG} in the era of LLMs, retrieval methods can be more personalized and advanced through search engines. \textbf{Hybrid Augmentation} has exceptional few-shot learning ability which is well-known in the prompt engineering era and includes a retrieval module that could retrieve additional knowledge, although standard RAG \cite{RAG} methods also have a prompt portion, we classify them as Retrieval-based Augmentation since it does not exhibit the few-shot learning capacity. The detailed data augmentation methods for four techniques in recent studies are depicted in Figure \ref{fig:DA_methods}.
\subsection{Simple Augmentation}
Simple Augmentation is an early data augmentation technique, such as synonym replacement, word deletion \cite{FlipDA} and back translation \cite{translation}. Many traditional data augmentation works have applied large language models to create diverse datasets and alleviate data scarcity issues.
\subsubsection{Text Transformation}
A conventional approach for data augmentation is through simple text transformations \cite{GPT3Mix}. GenAug \cite{GenAug} explores multiple text augmentation approaches, including randomly swapping two words' positions, deleting one word, or performing character insertion. For keyword replacement, this method uses RAKE \cite{RAKE} to extract keywords and incorporates external information through WordNet \cite{WordNet} by replacing keywords. The replaced keywords are sorted by their RAKE scores. One of DAGAM \cite{DAGAM} data augmentation schemes is to adopt a character order change strategy. This strategy involves fixing a word's first and last characters and randomly changing the remaining characters. MRC-QA \cite{MRC-QA} creates augmented data by shifting the correct answer span a few characters to the left or right. Each synthetic answer span is used as additional training data.

\subsubsection{Seed Selection Strategies}
During the model training phase, using selected seed training data could enable the model to learn key features of the data \cite{Human-in-the-loop}. Choosing a high-quality dataset for model training is crucial to reduce the number of training examples and lower computational costs \cite{AugmentedSBERT}. Selection-DA \cite{Human-in-the-loop} implements various seed selection strategies and highlights the impact of the seed training samples selection process on the quality of synthetic data and classifier performance. Further, it proposes a human-in-the-loop method for selecting seed samples.

\subsubsection{Back-translation}
Back-translation \cite{translation} is a typical data augmentation technique used in machine translation, which generates parallel training data by translating monolingual data from the target language back to the source language. AuGPT \cite{AuGPT} utilizes back-translation to paraphrase the original data to increase data diversity.

\subsubsection{Sequence-based Methods}
More recently, a growing number of researches have employed pre-trained language models to obtain augmented datasets. COCA \cite{COCA} randomly masks a fixed percentage of the input tokens to enhance the robustness of the user’s behaviour sentence representation and incorporates the user’s behaviour information into the document ranking process. TransformersDA \cite{TransformersDA} conditions the pre-trained models by prepending class labels in text sequences and produces one synthesized example for every example in the training set. LAMBADA \cite{LAMBADA} fine-tunes GPT-2 \cite{GPT-2} using existing labelled data. It leverages a simple form, using \texttt{[SEP]} and \texttt{[EOS]} to concatenate multiple sentences. For any class of labels $y$, use the fine-tuned language model to predict the duration of the sequence \texttt{y [SEP]} until \texttt{[EOS]}, allowing each class to synthesize any number of sentences. For the question-answering task, G-DAUGc \cite{G-DAUGc} fine-tunes a language model on the training question set $Q_i$ to train the question generator, where $Q_i$ is the word sequence representing the $i^{th}$ problem. Then, it generates new questions using the fine-tuned model and uses a similar method to create synthetic answers and distractors to obtain an augmented dataset. LeCA \cite{LeCA} designs the input representation to distinguish between the source sentence and each constraint and applies lexical constraints to augment data in neural machine translation.

\subsection{Prompt-based Augmentation}
					The success of prompt engineering has dramatically stimulated the capabilities of LLMs \cite{Psurvey}. Providing LLMs with carefully designed prompts can make LLMs produce more human-like responses. Prompt-based Augmentation technique effectively improve many downstream tasks' performance by providing task-related or cross-task prompts to guide LLMs in generating high-quality data.
					
     \subsubsection{Single-step Prompts}
Single-step Prompts only devise one-step instructions. The most common form is a straightforward question or a specific task requirement that does not require the model to perform multi-step reasoning.

					\textbf{Zero-shot Prompting} LLMs have been adjusted to follow instructions and are trained on the extensive corpus. Their enormous knowledge enables them to perform specific tasks in a zero-shot manner \cite{ZeroShotDataAug}. DA-NMT \cite{DA-NMT} provides storytelling, paraphrasing and multi-target prompts without any examples to acquire synthetic parallel datasets by utilizing ChatGPT. ZeroShotDataAug \cite{ZeroShotDataAug} employs LLMs to directly produce data based on the prompt in a zero-shot setting. 
                    
                    For information retrieval (IR) tasks, DAPDR \cite{DAPDR} creates relevant queries for a given text document in a zero-shot setting, then leverages IR models to acquire top-ranked candidate datasets for each synthetic query by calculating the cosine similarity between texts. UDAPDR \cite{UDAPDR} uses zero-shot prompting and also experiments with few-shot prompting to guide an LLM to create a set of synthesized queries. The queries generated from each prompt are applied to train a separate reranker, which is distilled into a single dense retriever for the target domain.

					\textbf{Few-shot Prompting} Unlike Zero-shot Prompting, Few-shot Prompting provides specific examples or one example to prompt LLMs to generate desired responses \cite{PE}. AugGPT \cite{AugGPT} uses ChatGPT \cite{GPT-3.5} as a data augmentation tool in few-shot learning. In particular, ChatGPT is employed to paraphrase each input sentence into a collection of extra sentences, thereby expanding the training samples. EPA \cite{EPA} and DAIL \cite{DAIL} apply a one-shot example to prompt the LLM to rephrase the original example and create new paraphrased examples. LLM–Assisted \cite{LLM–Assisted} leverages few-shot prompting to guide LLMs to paraphrase the original sentences or content. DA-intent \cite{DA-intent} addresses the difficulty of obtaining a large set of example utterances with the same intent by feeding an intent with available K-shot examples to an LLM. 
     
     In addition to common NLP tasks, many studies have exploited query generation in information retrieval by providing few-shot examples for prompt-based LLMs to improve retrieval performance significantly, such as InPars \cite{InPars} provides three examples of few-shot learning for the language model to generate document-related queries, then acquires the most likely optimal documents for each query. Promptagator \cite{Promptagator} directs LLMs with up to 8 examples in the prompt to produce a massive set of synthesized query-document pairs.

     \textbf{Cloze Prompting} involves filling in blanks in a sentence or context and operates similarly to masked language models known as BERT \cite{BERT}. Masking part of the input text and applying the model to infer the masked part based on the context. FlipDA \cite{FlipDA} forms a new sample by randomly masking a fixed percentage of the input contents and then employing a pre-trained language model to fill in the cloze blanks. SUNGEN \cite{PROMPTING} leverages prompts that are composed of \textit{\texttt{<mask>}} and \textit{\texttt{<mask>}} position will be replaced with the label word. GENIUS \cite{GENIUS} produces new samples based on the sketch of training samples. It fills in blanks \texttt{[M]} with several words or long spans while keeping the main parts of the sketch in the synthetic text, ensuring that the augmented text does not have significant semantic deviations from the initial text.

					\subsubsection{Multi-step Prompts}
To reduce the errors generated by the language model, unlike Single-step Prompts that ask the model to produce a direct response, Multi-step Prompts direct the generative model to provide a desired response step by step.
 
					\textbf{Chain-of-Thought Prompting} (CoT Prompting) guides LLMs to reason step by step to solve complex tasks \cite{PE}. A typical prompt contains a structure similar to \textit{let's think step by step.} LLM-PTM \cite{Healthcare-DA} proposes implementing Chain-of-Thought Prompting to direct LLMs in creating additional data points while maintaining semantic consistency in the inclusion and exclusion criteria of the original trials. LLM–Assisted \cite{LLM–Assisted} employs CoT Prompting to rewrite the source input content and gradually instruct LLMs for dependency tree generation. ConvAug \cite{ConvAug} introduces a three-step prompting approach motivated by the theory of human cognition to reduce hallucinations and improve data quality.
					
					\subsubsection{Structured Prompts}
     Structured Prompts explicitly provide the structure of the output content in the prompt template to adapt to more specific structured tasks. This ensures the model produces the expected content and complies with the standard framework.
    
    \textbf{Role Prompting} assigns specific roles or personas to the model, devising its response style and content based on predefined characteristics. AugESC \cite{AugESC} uses a task description and role prompts of \textit{Human} and \textit{AI} to distinguish between seekers and supporters. In the following dialogue completion process, the model receives the task description and a starting phrase starting with \textit{Human}. Then, it is fed with the next \textit{AI}. Finally, create the following dialogue. Inspired by LLM's self-criticism ability, ICLEF \cite{ICLEF} provides ChatGPT with a small amount of expert revision and asks it to behave as an annotator, criticizing its output results.

		 \textbf{Tuple Prompting} structures prompt as tuples or triples in a structured data pairs format to guide the model better understanding user’s intents \cite{PELLM}. GPT3Mix \cite{GPT3Mix} uses the triple consisted of Text Type \textit{T}, Label Type \textit{L}, and Label-token Verbalizer \textit{v: Y $\rightarrow$ V} to create the task specification \textit{S = (T, L, v)}. Each task has a task description and is used to construct prompts.

\textbf{Template Prompting} utilizes structured templates that guide the model to respond according to a designed format or instruction. LLM-DA \cite{LLM-DA} provides LLMs with given sentences and entities and requires LLMs to construct the expected format. After giving the definitions of  \textit{age\_limit} and  \textit{atm\_support}, PromptMix \cite{PromptMix} requires the LLMs to produce utterances composed of different proportions of the  \textit{age\_limit} and  \textit{atm\_support}. Unnatural-instructions \cite{Unnatural-instructions} constructs well-defined input formats and applies them to the article summarization task. LLM2LLM \cite{LLM2LLM} gives detailed requirements in the prompt and asks the model to return a structured response. Other studies, such as WANLI \cite{WANLI} and LLM-powered \cite{ LLM-powered}, also adopt a predefined template to instruct LLMs to generate new data.

%% file: Table/Task.tex
\begin{table*}[htbp]
\caption{Data Augmentation tasks, sub-tasks, and datasets.}
\label{DA task, sub-task, dataset}
             \centering
          \scalebox{0.85}{  
\begin{tabular}{llll}
\toprule
Tasks                                       & Sub-tasks                                  & Datasets                                                                                                          & Methods                                                                                                                                                                                                                                                                                 \\
\midrule
\multirow{12}{*}{Text Classification}       & \multirow{5}{*}{Sentiment Classification}  & CR, Yelp, Yelp5, Tweet-Eval, MR                                                                                   & zicl  \cite{zicl}                                                                                                                                                                                                                                                                                                                                                                         \\
                                            &                                            & SST-2                                                                                                             & \begin{tabular}[c]{@{}l@{}}TransformersDA  \cite{TransformersDA},GPT3Mix  \cite{GPT3Mix},LLM2LLM  \cite{LLM2LLM},\\ DAIL  \cite{DAIL}, ZeroShotDataAug  \cite{ZeroShotDataAug},\\ SUNGEN  \cite{PROMPTING},GENIUS  \cite{GENIUS}, zicl  \cite{zicl}\end{tabular} \\
                                            &                                            & IMDb                                                                                                              & DAGAM  \cite{DAGAM},SUNGEN  \cite{PROMPTING},GENIUS  \cite{GENIUS}                                                                                                                                                                                                                                                                                      \\
                                            &                                            & Amazon                                                                                                            & AugGPT  \cite{AugGPT},SUNGEN  \cite{PROMPTING}, DAICL \cite{DAICL}, zicl  \cite{zicl}                                                                                                                                                                                                                                                  \\
                                            &                                            & Twitter Complaints                                                                                                & PromptMix  \cite{PromptMix}                                                                                                                                                                                                                                                                                                                                                               \\
                                            \cline{2-4}
                                            & \multirow{2}{*}{Intent Classification}     & SNIPS                                                                                                             & \begin{tabular}[c]{@{}l@{}}TransformersDA  \cite{TransformersDA},DA-intent  \cite{DA-intent},\\ LLM2LLM  \cite{LLM2LLM},ZeroShotDataAug  \cite{ZeroShotDataAug}\end{tabular}                                                                                                                                                           \\
                                            &                                            & Banking                                                                                                           & DA-intent  \cite{DA-intent},PromptMix  \cite{PromptMix}                                                                                                                                                                                                                                                                                                                  \\
                                            \cline{2-4}
                                            & \multirow{2}{*}{Question Classification}   & TREC                                                                                                              & \begin{tabular}[c]{@{}l@{}}TransformersDA  \cite{TransformersDA},GPT3Mix  \cite{GPT3Mix},\\ DAGAM  \cite{DAGAM},LLM2LLM  \cite{LLM2LLM}, LAMBADA  \cite{ LAMBADA}\\ ZeroShotDataAug  \cite{ZeroShotDataAug},PromptMix  \cite{PromptMix}\end{tabular}                              \\
                                            &                                            & Quora-QP                                                                                                          & AugmentedSBERT \cite{AugmentedSBERT}                                                                                                                                                                                                                                                                                                                                                      \\
                                            \cline{2-4}
                                            & \multirow{2}{*}{Topic Classification}      & AGnews                                                                                                            & DAGAM  \cite{DAGAM},DAIL \cite{DAIL},SUNGEN \cite{PROMPTING}                                                                                                                                                                                                                                                                                            \\
                                            &                                            & Yahoo                                                                                                             & DAIL \cite{DAIL},GENIUS \cite{GENIUS}                                                                                                                                                                                                                                                                                                                                    \\
                                            \cline{2-4}
                                            & Subjectivity Classification                & Subjectivity                                                                                                      & SUNGEN \cite{PROMPTING},PromptMix \cite{PromptMix}                                                                                                                                                                                                                                                                                                                       \\
                    \hline

\multirow{19}{*}{Text Generation}           & Reviews Generation                         & Yelp Reviews                                                                                                      & GenAug \cite{GenAug}                                                                                                                                                                                                                                                   \\
\cline{2-4}
                                            & \multirow{3}{*}{Machine Translation}       & AI-hub corpus                                                                                                     & DA-NMT \cite{DA-NMT}                                                                                                                                                                                                                                                                                                                                                                      \\
                                            &                                            & newstest2014                                                                                                      & LeCA \cite{LeCA}                                                                                                                                                                                                                                                                                                                                                                          \\
                                            &                                            & FLORES-200                                                                                                        & EPA \cite{EPA}                                                                                                                                                                                                                                                                                                                                                                            \\
                                        \cline{2-4}
                                            & Paraphrasing                               & Quora Question Pairs (QQP)                                                                                        & EPA \cite{EPA}                                                                                                                                                                                                                                                                                                                                                                            \\
                                            \cline{2-4}
                                            & \multirow{12}{*}{Dialogue Generation}      & Reddit                                                                                                            & CGRG \cite{CGRG}, RetGen \cite{RetGen}                                                                                                                                                                                                                                                                                                                                   \\
                                            &                                            & Dailydialog, fraudulent e-mails                                                                                   & EDGE \cite{EDGE}                                                                                                                                                                                                                                                                                                                                                                          \\
                                            &                                            & DuLeMon,KBP                                                                                                       & UniMS-RAG \cite{UniMS-RAG}                                                                                                                                                                                                                                                                                                                                                                \\
                                            &                                            & Wizard of the Internet                                                                                            & Internet-Aug \cite{Internet-Aug}                                                                                                                                                                                                                                                                                                                                                          \\
                                            &                                            & Wizard-of-Wikipedia                                                                                               & DialogGen \cite{DialogGen}                                                                                                                                                                                                                                                                                                                                                                \\
                                            &                                            & \begin{tabular}[c]{@{}l@{}}Common Document Corpus, Social Media Data,\\ Benchmark Tasks\end{tabular}              & ChatPLUG \cite{ChatPLUG}                                                                                                                                                                                                                                                                                                                                                                  \\
                                            &                                            & MultiWOZ 2.1                                                                                                      & Efficient-RAG  \cite{Efficient-RAG}                                                                                                                                                                                                                                                                                                                                                       \\
                                            &                                            & CONVAI2                                                                                                           & LAPDOG  \cite{LAPDOG}                                                                                                                                                                                                                                                                                                                                                                     \\
                                            &                                            & ESConv                                                                                                            & AugESC  \cite{AugESC}                                                                                                                                                                                                                                                                                                                                                                     \\
                                            &                                            & MultiWOZ                                                                                                          & AuGPT  \cite{AuGPT}                                                                                                                                                                                                                                                                                                                                                                       \\
                                            &                                            & \begin{tabular}[c]{@{}l@{}}PersonaChat,Empathetic Dialogues,\\ Blended Skill Talk,Multi-Session Chat\end{tabular} & SeeKeR  \cite{SeeKeR}                                                                                                                                                                                                                                                                                                                                                                     \\
                                            &                                            & Schema Guided Dialog                                                                                              & Synthetic-DA  \cite{Synthetic-DA}                                                                                                                                                                                                                                                                                                                                                         \\
                                            \cline{2-4}
                                            & \multirow{2}{*}{Dialogue Summarization}    & MIMIC-III MTSamples                                                                                               & Dialogue-Convert  \cite{Dialogue-Convert}                                                                                                                                                                                                                                                                                                                                                 \\
                                            &                                            & SAMSum                                                                                                            & EPA  \cite{EPA}                                                                                                                                                                                                                                                                                                                                                                           \\
                                               \hline
\multirow{6}{*}{Information Extraction}     & \multirow{3}{*}{Event Argument Extraction} & WikiEvents                                                                                                        & EAE-RAG  \cite{EAE-RAG}, RGQA  \cite{ RGQA}                                                                                                                                                                                                                                                                                                                              \\
                                            &                                            & ACE 2005                                                                                                          & X-GEAR  \cite{X-GEAR}, RGQA  \cite{RGQA}                                                                                                                                                                                                                                                                                                                                 \\
                                            &                                            & RAMS                                                                                                              & EAE-RAG  \cite{EAE-RAG}                                                                                                                                                                                                                                                                                                                                                                   \\
                                            \cline{2-4}
                                            & \multirow{3}{*}{Named Entity Recognition}  & OntoNotes 5.0,MIT-Movie,FEW-NERD                                                                                  & LLM-DA  \cite{LLM-DA}                                                                                                                                                                                                                                                                                                                                                                     \\
                                            &                                            & CoNLL03                                                                                                           & GENIUS  \cite{GENIUS}, LLM-DA  \cite{LLM-DA}, DAICL  \cite{DAICL}                                                                                                                                                                                                                                                                                       \\
                                            &                                            & WNUT, FIN, BC2GM , BioNLP09, BC5CDR                                                                               & DAICL  \cite{DAICL}                                                                                                                                                                                                                                                                                                                                                                       \\
                                               \hline

\multirow{14}{*}{Question Answering}        & \multirow{8}{*}{Single-hop}                & TechQA                                                                                                            & MRC-QA \cite{MRC-QA}, Read-Com  \cite{Read-Com}, RADA  \cite{Read-Com}                                                                                                                                                                                                                                                                                  \\
                                            &                                            & PolicyQA                                                                                                          & MRC-QA \cite{MRC-QA}, Read-Com  \cite{Read-Com}, RADA  \cite{Read-Com}                                                                                                                                                                                                                                                                                  \\
                                            &                                            & CovidQA                                                                                                           & Read-Com  \cite{Read-Com}, RADA  \cite{RADA}                                                                                                                                                                                                                                                                                                                             \\
                                            &                                            & BoolQ                                                                                                             & FlipDA \cite{FlipDA}                                                                                                                                                                                                                                                                                                                                                                      \\
                                            &                                            & SQUAD                                                                                                             & Generative-DA  \cite{Generative-DA}, GENIUS  \cite{GENIUS}                                                                                                                                                                                                                                                                                                               \\
                                            &                                            & NewWiki,NYT,Amazon                                                                                                & Generative-DA  \cite{Generative-DA}                                                                                                                                                                                                                                                                                                                                                       \\
                                            &                                            & WebQuestionsSP                                                                                                    & KAPING  \cite{KAPING}                                                                                                                                                                                                                                                                                                                                                                     \\
                                            &                                            & Natural Questions                                                                                                 & QA-Internet  \cite{QA-Internet}                                                                                                                                                                                                                                                                                                                                                           \\
                                            \cline{2-4}
                                            & \multirow{6}{*}{Multi-hop QA}              & ASQA, QAMPARI, ELI5                                                                                               & ALCE \cite{ALCE}                                                                                                                                                                                                                                                                                                                                                                          \\
                                            &                                            & Reddit                                                                                                            & Generative-DA  \cite{Generative-DA}                                                                                                                                                                                                                                                                                                                                                       \\
                                            &                                            & Mintaka                                                                                                           & KAPING  \cite{KAPING}                                                                                                                                                                                                                                                                                                                                                                     \\
                                            &                                            & HotPotQA                                                                                                          & IM-RAG  \cite{IM-RAG}, ReAct  \cite{ReAct}, QA-Internet  \cite{QA-Internet}                                                                                                                                                                                                                                                                            \\
                                            &                                            & STRATEGYQA                                                                                                        & QA-Internet  \cite{QA-Internet}                                                                                                                                                                                                                                                                                                                                                           \\
                                            &                                            & MultiRC, ReCoRD                                                                                                   & FlipDA \cite{FlipDA}                                                                                                                                                                                                                                                                                                                                                                      \\
                                               \hline
            \multirow{8}{*}{Natural Language Inference} & Multi-task NLI                             & MultiNLI                                                                                                          & WANLI \cite{WANLI}, EPA \cite{EPA}                                                                                                                                                                                                                                                                                                                                       \\
                                            & Multilingual Commonsense Reasoning         & XCOPA, XWinograd, XStoryCloze                                                                                     & LLM-powered \cite{LLM-powered}                                                                                                                                                                                                                                                                                                                                                            \\
                                            & Multiple-choice Reasoning                  & GSM8K, CaseHOLD                                                                                                   & LLM2LLM \cite{LLM2LLM}                                                                                                                                                                                                                                                                                                                                                                    \\
                                            & Textual Entailment                         & SNLI                                                                                                              & EPA \cite{EPA}, G-DAUGc \cite{G-DAUGc}                                                                                                                                                                                                                                                                                                                                   \\
                                            & Word Relation Reasoning                    & SUPERNI                                                                                                           & TAPP \cite{Prefix-Prompt}                                                                                                                                                                                                                                                                                                                                                                 \\
                                            & Fact Verification                          & FEVER                                                                                                             & QA-Internet  \cite{QA-Internet}                                                                                                                                                                                                                                                                                                                                                           \\
                                            & \multirow{2}{*}{Commonsense Reasoning}     & \begin{tabular}[c]{@{}l@{}}COMMONSENSEQA, WINOGRANDE,\\  CODAH and HellaSwag\end{tabular}                         & G-DAUGc \cite{G-DAUGc}                                                                                                                                                                                                                                                                                                                                                                    \\
                                            &                                            & Semeval-2012                                                                                                      & FlipDA \cite{FlipDA}                                                                                                                                                                                                                                                                                                                                                                      \\
                                               \hline
\multirow{6}{*}{Information Retrieval}      &                                            & \begin{tabular}[c]{@{}l@{}}MARCO, TREC-DL, Robust04, \\ Natural Questions, TREC-COVID\end{tabular}                & InPars \cite{InPars}                                                                                                                                                                                                                                                                                                                                                                      \\
                                            &                                            & BEIR (exclude MS MARCO, NQ and Quora)                                                                             & Promptagator \cite{Promptagator}                                                                                                                                                                                                                                                                                                                                                          \\
                                            &                                            & QReCC, TopiOCQA, CAsT-20 and CAsT-21                                                                              & ConvAug \cite{ConvAug}                                                                                                                                                                                                                                                                                                                                                                    \\
                                            &                                            & NTCIR and ACODAR                                                                                                  & DAPDR \cite{DAPDR}                                                                                                                                                                                                                                                                                                                                                                        \\
                                            &                                            & AOL search log and Tiangong-ST                                                                                    & COCA \cite{COCA}                                                                                                                                                                                                                                                                                                                                                                          \\
                                            &                                            & se LoTTE, BEIR, NQ, and SQuAD                                                                                     & UDAPDR \cite{UDAPDR}                                                                                                                                                                                                                                                                                                                                                                      \\
                                               \hline
Regression                                  &                                            & STS,BWS                                                                                                           & AugmentedSBERT \cite{AugmentedSBERT}     \\
								\bottomrule
								
							\end{tabular}
							
 }
						\end{table*}

%% file: sec-Rmethods.tex
\subsection{Retrieval-based Augmentation}

Although LLMs have shown satisfactory capabilities in many fields, they inevitably suffer from producing hallucinations and being unable to use external information \cite{QA-Internet}. Retrieval-based Augmentation effectively overcomes some existing limitations of LLMs and provides an innovative way for data augmentation by retrieving enormous and dynamic knowledge from corpus bases or external documents \cite{IM-RAG}. Recently, many studies have used Retrieval-augmented Generation (RAG) \cite{RAG} to obtain external timely updated information and achieved exceptional performance on different tasks. Retrieval-based Augmentation methods could be categorised into Sparse Retrieval, Dense Retrieval, Graph-based Retrieval and Search Engine Retrieval.

\subsubsection{Sparse Retrieval}	
Sparse Retrieval is a traditional information retrieval method that relies on explicit word matching in documents. The most commonly used models include TF-IDF \cite{TF-IDF} and BM25 \cite{BM25}.

\textbf{TF-IDF} \cite{TF-IDF} computes text relevance by calculating the word’s frequency weights in the document and the whole corpus \cite{IFsurvey}. CGRG \cite{CGRG} employs IDF-weighted to rank grounding information from external knowledge sources to allow the model to create answers consistent with the facts. The main idea of using IDF-weighted is to measure word overlaps and select words with high overlaps and IDF weights. 

\textbf{BM25} \cite{BM25} is an improved version of TF-IDF, which introduces a document length normalisation factor \cite{IFsurvey}. AugmentedSBERT \cite{AugmentedSBERT} uses BM25 to sample similar sentences from unlabeled sentence pairs to form a silver dataset and then merge the silver and gold datasets for prediction.

\subsubsection{Dense Retrieval}
Dense Retrieval maps the query and documents into the same continuous vector space and calculates the distance between the vectors to measure the similarity. Compared to Sparse Retrieval, Dense Retrieval is superior at capturing semantic information \cite{IFsurvey}.

\textbf{DPR} \cite{DPR} uses a dual-tower structure to encode queries and documents separately. It obtains semantic representation based on BERT structure and optimises the model through comparative learning \cite{DPR}. IM-RAG \cite{IM-RAG} implements DPR to embed documents and then uses the FAISS library \cite{FAISS} for fast semantic similarity retrieval.

\textbf{ANCE} \cite{ANCE} is a dense retrieval model that uses an asynchronously updated pool of negative samples for contrastive learning, which improves the ability to distinguish negative samples \cite{ANCE}. RetGen \cite{RetGen} adopts ANCE as the dense retrieval model and initialises the dense retriever with ANCE.

\textbf{SimCSE} \cite{SimCSE} is a sentence embedding model trained by contrast learning. It improves the semantic representation of sentences by maximising the similarity of positive sample pairs and minimising the similarity of negative sample pairs using both labelled and unlabelled data \cite{SimCSE}. zicl \cite{zicl} proposes to retrieve sentences with similar input distributions to the test input by calculating the cosine similarity between two different sentence embeddings using SimCSE.

\textbf{Poly-encoder} \cite{Poly-encoder} captures the multilayer semantics of an input text by using multiple fixed numbers of context vectors, which can efficiently represent different information in long texts \cite{Poly-encoder}. EDGE \cite{EDGE} implements a poly-encoder model to perform retrieval and then choose the highest-ranked exemplar responses.

\textbf{S-BERT} \cite{S-BERT} is a BERT-based model specialised for sentence embedding generation. It captures the semantic similarity between sentences by applying a dual-encoder architecture \cite{S-BERT}. Compared with traditional BERT, S-BERT significantly improves the efficiency of similarity computation tasks. EAE-RAG \cite{EAE-RAG} leverages S-BERT to retrieve top-K ranked documents with the highest relevance to the original input document from the training corpus and uses the retrieved documents as a set of discrete demonstrations. RGQA \cite{RGQA} first introduces three additional tokens \texttt{[demo]}, \texttt{[tgr]}, and \texttt{[sep\_arg]} to obtain distinct parts of the final input sequence. \texttt{{[}demo{]}} represents the most significant demonstration of the question and input examples. Then, it utilizes S-BERT and computes the similarity scores to retrieve the demonstration.

\textbf{Contriever} \cite{Contriever} is a retrieval model pre-trained with unsupervised data using a dual encoder architecture. It encodes query and document independently and obtains the similarity between query and document usually by calculating the dot product of their corresponding outputs \cite{Contriever}. LAPDOG \cite{LAPDOG} implements Contriever as the retriever to embed query and story corpus separately. Then, the retriever computes the dot product similarity between the query and each story to obtain stories with the top-K ranked similarity scores.

\textbf{RoBERTa} \cite{RoBERTa} could be used as an encoder to encode queries and documents. It converts text into vector representations and calculates the similarity between these vectors to find the most relevant documents to the query from a large-scale document collection. Efficient-RAG \cite{Efficient-RAG} proposes to utilize a siamese network structure with two encoders to encode dialogue context and knowledge snippets separately and establish a suitable ranking function using the distance between them. It uses pre-trained RoBERTa \cite{RoBERTa} models as the encoders.

\textbf{TAS-B} \cite{TAS-B} combines BERT dense representation and sparse features to optimise long text retrieval through multi-level semantic aggregation and contrastive learning \cite{TAS-B}. Detailed methods will be introduced in Section \ref{Single-step Prompts and Dense Retrieval}.

\textbf{GTR} \cite{GTR} is a dense retrieval method based on the T5 \cite{T5}. It converts queries and documents into dense vectors and identifies candidate documents by calculating the similarity to achieve efficient information retrieval \cite{GTR}. In Section \ref{Single-step Prompts and Dense Retrieval}, we will introduce how the recent paper uses GTR for retrieval.

\subsubsection{Graph-based Retrieval}
Graph-based Retrieval typically utilizes graph structures to retrieve documents and queries. To improve LLM’s performance, Personae-DA \cite{Personae-DA} introduces a new memory framework in the information retrieval model to reflect how humans dynamically access memory during interactive processes and use a memory model with an adaptive graph-based architecture to obtain relevant retrieved data. 

\subsubsection{Search Engine Retrieval}
Search Engine Retrieval is a technique for retrieval-related tasks through external knowledge bases. The model's output may be out of control or inaccurate without providing external knowledge \cite{CGRG}. Retrieving external documents and injecting grounding knowledge into the model helps the model generate faithful and context-aware responses.

\textbf{Wikipedia Search} is a system designed for searching within the Wikipedia database. DialogGen \cite{DialogGen} employs the query generator to generate queries based on the conversation context. This method then implements Wikipedia Search \footnote{\url{https://en.wikipedia.org/wiki/Special:Search}} as the search engine to acquire significant articles. The response generator generates responses based on the retrieved articles and the conversation context.

\textbf{APIs} return search results by calling interfaces provided by a particular platform or service (e.g. Google, Bing, and Wikipedia). ChatPLUG \cite{ChatPLUG} accesses the internet-relevant knowledge by Quark Search API \footnote{\url{https://quark.sm.cn/}} based on the conversation context. Internet-Aug \cite{Internet-Aug} first employs the Bing Search API to generate a URL list for each query, then retrieves up-to-the-minute relevant information based on these URLs, and the retrieved knowledge is prepended to the conversation history. ReAct \cite{ReAct} performs retrieval by the Wikipedia web API to search for an entity and return the first five sentences of the corresponding entity's wiki page if it exists. Given the synthesised query, SeeKeR \cite{SeeKeR} adopts the Bing Web Search API \footnote{\url{www.microsoft.com/en-us/bing/apis/bing-web-search-api}} to retrieve documents. The retrieved document is used to produce a knowledge response composed of relevant sentences or phrases.

\subsection{Hybrid Augmentation}		
The combination of constructing prompts and retrieval components not only stimulates LLMs to produce diverse datasets but also exploits retrievers to obtain timely updated information. Hybrid Augmentation technique is shown in a grey font in Figure \ref{fig:DA_methods}. There could be many different combinations to construct prompts and perform retrieval. The following will introduce the various combinations of the papers covered in this survey.

\subsubsection{Single-step Prompts and Sparse Retrieval}
\label{Single-step Prompts and Sparse Retrieval}
UniMS-RAG \cite{UniMS-RAG} adopts existing methods such as TF-IDF and BM25 to obtain similarity scores between dialogue context and evidence. Motivated by recent studies that use LLMs to predict similarity relationships between queries and evidence, it provides dialogue context and multiple pieces of evidence to prompt ChatGPT \cite{GPT-3.5} to predict similarity scores in a zero-shot setting. ALCE \cite{ALCE} employs BM25 to retrieve relevant passages from an extensive collection of documents and returns candidate passages. It constructs two in-context demonstrations in input prompts and applies them to an LLM to interact with the sparse retriever.

\subsubsection{Single-step Prompts and Dense Retrieval}
\label{Single-step Prompts and Dense Retrieval}
For the sentiment analysis task, DAICL \cite{DAICL} leverages SimCSE Roberta as the retrieval model to retrieve examples semantically similar to the input sentence from the target unlabeled corpus as the context of the source query. Then, it concatenates the source query and the retrieved context as input few-shot demonstrations for domain-adaptive in-context learning. To acquire examples similar to the original samples, RADA \cite{RADA} implements DistilBert TAS-B to retrieve examples from external resources and leverages the retrieved samples to create new input-output pairs in the few-shot in-context setting. Given a question, ALCE \cite{ALCE} first employs GTR \cite{GTR} and DPR \cite{DPR} to generate content while providing related cited passages from a vast retrieval corpus. Followed by prompting an LLM to create and cite related evidence without fine-tuning the model’s internal parameters. 

\subsubsection{Single-step Prompts and Search Engine Retrieval}
QA-Internet \cite{QA-Internet} applies the question verbatim as a query and leverages Google Search API \footnote{\url{https://developers.google.com/custom-search}} to extend knowledge of the model by retrieving a set of documents for each query. Followed by adopting k-shot prompting and the retrieved paragraphs for question-answering scenarios.

\subsubsection{Multi-step Prompts and Search Engine Retrieval}
ReAct \cite{ReAct} randomly selects cases from the training set and manually constructs trajectories that compose diverse thought-action-observation processes. It performs retrieval by the Wikipedia web API to search for an entity and return the first five sentences of the corresponding entity's wiki page if it exists.

\subsubsection{Structured Prompts and Graph-based Retrieval}
\label{Structured Prompts and Graph-based Retrieval}
 KAPING \cite{KAPING} uses MPNet \cite{MPNet} as the symmetric retriever and applies TAS-B \cite{TAS-B} as the asymmetric retriever to retrieve only the related facts from the external Knowledge Graph. Then, it augments them to the prompt composed of relevant fact triples to produce a grounded response.

%% file: sec-Post-processing.tex
The augmented data by LLMs is not fully guaranteed to be valid for training, and it is crucial to refine the generated data and further ensure data quality and relevance \cite{LLM-powered}. We categorize the post-processing approaches into consistency measures, filtering techniques, heuristic methods, and human involvement. 

 \subsection{Consistency Measures}
Consistency measures ensure that the generated data is logically and semantically consistent with the original data. LAMBADA \cite{LAMBADA} ranks the generated sentences for the same label by calculating the confidence score and keeping the top-ranked sentences. After developing a set of new examples, LLM-powered \cite{LLM-powered} performs post-processing by only adding efficient and consistent output samples to the training set. Promptagator \cite{Promptagator} and Generative-DA \cite{Generative-DA} clean the synthetic data by employing round-trip consistency. Many studies introduce multiple reranking strategies to refine the response. ALCE \cite{ALCE} utilizes the citation recall store as a rerank strategy to further improve the output's quality. Aiming to get the top-ranked supporting knowledge,  IM-RAG \cite{IM-RAG} leverages a refiner to retain significant information of the retriever's output.

 \subsection{Filtering Techniques}
Filtering techniques are used to retain generated data that is helpful to the model and to filter redundant or invalid data. G-DAUGc \cite{G-DAUGc} applies influence functions to filter out detrimental training examples. LLM-DA \cite{LLM-DA}, Unnatural-instructions \cite{Unnatural-instructions} and RGQA \cite{RGQA} remove non-qualified augmented content by implementing specific filtering metrics. To select high-quality synthetic queries, UDAPDR \cite{UDAPDR} applies a filter to refine generated queries that could return its gold passage among the first 20 of the retrieved results. To further enhance the quality of generated data, Read-Com \cite{Read-Com} harnesses a round-trip filtration strategy. LLM2LLM \cite{LLM2LLM} and DAPDR \cite{DAPDR} harness similarity-based filtering approaches, which specifically use ROUGE filter and compute cosine similarity, respectively. Aiming to ensure the LLM could generate examples for the majority class, PromptMix \cite{PromptMix} selects top-5 classes based on the similarity between the synthetic examples and current examples. MRC-QA \cite{MRC-QA} only retains the documents related to answer spans that obtain the top-K highest scores. To reduce the number of retrieved factual triples, KAPING \cite{KAPING} only keeps the top-K relevant triples to the question.

 \subsection{Heuristic Methods}
Heuristic methods avoid common problems by adopting specific rules or principles. AugESC \cite{AugESC} discards undesirable generated content based on designed heuristics in the last post-processing stage. WANLI \cite{WANLI} firstly filters failure examples by applying heuristics and then calculates the estimated max variability for the rest of the examples.

\subsection{Human Involvement}
Human revision plays a critical role in refining augmented datasets. WANLI \cite{WANLI} recruits crowd-workers to review unlabeled examples and remove offensive examples.

The practical application of post-processing approaches improves the quality of the augmented datasets and enhances the overall performance of language models \cite{UDAPDR}, making them more robust and reliable.

%% file: Table/Evaluation.tex
\begin{table*}[!h]
							\caption{Evaluation Metrics of four Data Augmentation Techniques.}
       \label{Evaluation}
       \centering
             \scalebox{1.0}{   
							\begin{tabular}{llll}
								\toprule
								DA techniques                            & Evaluation                            & Metrics            & Methods                                                                                                                                                        \\
								\midrule
\multirow{6}{*}{Simple Augmentation} & \multirow{6}{*}{Automatic Evaluation} & Accuracy          & DAGAM \cite{DAGAM}, MRC-QA \cite{MRC-QA} \\
                                     &                                       & Exact Match score & MRC-QA \cite{MRC-QA}                 \\
                                     &                                       & Recall            & AuGPT \cite{AuGPT}, MRC-QA \cite{MRC-QA}        \\
                                     &                                       & F1 Score          & AuGPT \cite{AuGPT}, Selection-DA \cite{Human-in-the-loop}, MRC-QA \cite{MRC-QA}  \\
                                     &                                       & Perplexity        & GenAug \cite{GenAug}                     \\
                                     &                                       & BLEU              & GenAug \cite{GenAug}, AuGPT \cite{AuGPT}, LeCA \cite{LeCA}        \\
\hline
								\multirow{11}{*}{Prompt-based Augmentation}       & \multirow{7}{*}{Automatic Evaluation} & Accuracy          & \begin{tabular}[c]{@{}l@{}}WANLI \cite{WANLI},LLM2LLM \cite{LLM2LLM},FlipDA \cite{FlipDA},GPT3Mix \cite{GPT3Mix},DA-intent \cite{DA-intent}\\ DAIL \cite{DAIL},EPA \cite{EPA},SUNGEN \cite{PROMPTING},PromptMix \cite{PromptMix}\end{tabular}                                           \\
								&                                       & ROUGE             & Dialogue-Convert \cite{Dialogue-Convert},Synthetic-DA \cite{Synthetic-DA},EPA \cite{EPA},TAPP \cite{Prefix-Prompt}                                                                                                              \\
								&                                       & Exact Match score & Read-Com \cite{Read-Com},TAPP \cite{Prefix-Prompt},Generative-DA \cite{Generative-DA}                                                                                                                         \\
								&                                       & Recall            & DA-intent \cite{DA-intent},GENIUS \cite{GENIUS}                                                                                                                                             \\
								&                                       & F1 Score          & \begin{tabular}[c]{@{}l@{}}Dialogue-Convert \cite{Dialogue-Convert},LLM-DA \cite{LLM-DA},X-GEAR \cite{X-GEAR},Read-Com \cite{Read-Com}\\ Generative-DA \cite{Generative-DA},EPA \cite{EPA},FlipDA \cite{FlipDA}\end{tabular}                                  \\
								&                                       & Perplexity        & GENIUS \cite{GENIUS}                                                                                                                                                        \\
								&                                       & BLEU              & DA-NMT \cite{DA-NMT},EPA \cite{EPA}                                                                                                                                                    \\
					\cline{2-4}
								& \multirow{4}{*}{Human Evaluation}     & Consistency       & \multirow{4}{*}{AugESC \cite{AugESC}}                                                                                                                                       \\
								&                                       & Coherence         &                                                                                                                                                               \\
								&                                       & Informativeness   &                                                                                                                                                               \\
								&                                       & Safety            &                                                                                                                                                               \\
	\hline
								\multirow{15}{*}{Retrieval-based Augmentation}    & \multirow{7}{*}{Automatic Evaluation} & Accuracy          & zicl \cite{zicl}                                                                                                                             \\
								&                                       & ROUGE             & LAPDOG \cite{LAPDOG}                                                                                                                      \\
								&                                       & Exact Match score &IM-RAG\cite{IM-RAG}                                                                                                                                                 \\
								&                                       & Recall            & DialogGen \cite{DialogGen}                                                                                                                                             \\
								&                                       & F1 Score          & \begin{tabular}[c]{@{}l@{}}EAE-RAG \cite{EAE-RAG},X-GEAR \cite{X-GEAR},IM-RAG \cite{IM-RAG},AugmentedSBERT \cite{AugmentedSBERT}\\ Internet-Aug \cite{Internet-Aug},DialogGen \cite{DialogGen},Efficient-RAG \cite{Efficient-RAG},LAPDOG \cite{LAPDOG}\\\end{tabular} \\
								&                                       & Perplexity        & Internet-Aug \cite{Internet-Aug},DialogGen \cite{DialogGen}                                                                                                                  \\
								&                                       & BLEU              & RetGen \cite{RetGen},Efficient-RAG \cite{Efficient-RAG},LAPDOG \cite{LAPDOG}                                      \\
		\cline{2-4}
								& \multirow{8}{*}{Human Evaluation}     & Consistency       & SeeKeR \cite{SeeKeR}                                                   \\
								&                                       & Coherence         & ChatPLUG \cite{ChatPLUG},RetGen \cite{RetGen}                                                                                                                                           \\
								&                                       & Informativeness   & ChatPLUG \cite{ChatPLUG},RetGen \cite{RetGen}                                                                                                                                                     \\
								&                                       & Safety            & ChatPLUG \cite{ChatPLUG}                                                                                                                                                       \\
								&                                       & Hallucination     & ChatPLUG \cite{ChatPLUG}                                                                                                                                       \\
								&                                       & Knowledgeable     & Internet-Aug \cite{Internet-Aug}, SeeKeR \cite{SeeKeR}                                                                                                                                    \\
								&                                       & Engaging          & Internet-Aug \cite{Internet-Aug}                                                                                                                       \\
	\hline
								\multirow{10}{*}{Hybrid Augmentation} & \multirow{6}{*}{Automatic Evaluation} & Accuracy          & KAPING \cite{KAPING},ReAct\cite{ReAct},QA-Internet \cite{QA-Internet},DAICL \cite{DAICL}                                                                                                                                \\
                                &                                        &   ROUGE         & UniMS-RAG\cite{UniMS-RAG}    \\
								&                                       & Exact Match score & RGQA \cite{RGQA},ReAct \cite{ReAct},QA-Internet \cite{QA-Internet}                                                                                                                                        \\
								&                                       & Recall            & RGQA \cite{RGQA},ConvAug \cite{ConvAug}                                                                                                                                                  \\
								&                                       & F1 Score          & RGQA \cite{RGQA},RADA \cite{RADA},DAICL \cite{DAICL},UniMS-RAG \cite{UniMS-RAG}                                                                                                                                               \\                                                                                                             
        						&                                       & BLEU          & CGRG \cite{CGRG},UniMS-RAG \cite{UniMS-RAG}                                                                                                                                           \\
		\cline{2-4}
								& \multirow{4}{*}{Human Evaluation}     & Consistency       & SeeKeR \cite{SeeKeR}                                                                                                                          \\
    								&                                       & Coherence    & UniMS-RAG \cite{UniMS-RAG}                                                                                                                                             \\
								&                                       & Fluency    & ALCE \cite{ALCE}                                                                                                                                            \\
								&                                       & Factually Correct & ALCE \cite{ALCE}  \\
								\bottomrule 
								
							\end{tabular}
	}
						\end{table*}
						

%% file: sec-applications.tex
\subsection{Tasks}
More recently, the parameters of deep neural network models have significantly increased, and the quantity and quality of data have become an essential role in the model's training process and received significant attention. Large-scale language models alleviate data scarcity by generating more augmented datasets with similar distribution to the original data \cite{ABSA}. Various data augmentation techniques extend existing text data and significantly improve the performance of natural language processing (NLP) tasks. Common NLP tasks include Information Extraction, Question Answering, and Text Classification \cite{DL-ABSA}.
						
						 In addition to typical NLP tasks, LLMs have been employed to improve the accuracy and new domain adaptability of information retrieval (IR) tasks \cite{UDAPDR}. In information retrieval, given a query, candidate documents related to the query are retrieved in the document set. A widely used method in the context of LLMs for IR tasks is query generation for data augmentation, which applies LLMs to generate queries or query-document pairs \cite{DAPDR}. The synthetic queries could help the retrieval model adapt to targeted domains in multiple tasks. InPars \cite{InPars}, Promptagator \cite{Promptagator}, UDAPDR \cite{UDAPDR} and DAPDR \cite{DAPDR} improve the performance of passage and document retrieval tasks by enhancing queries. Besides, COCA \cite{COCA} and  ConvAug \cite{ConvAug} utilize contrastive learning to obtain augmented data in document ranking and conversational dense retrieval tasks, respectively. 
       
       Moreover, With the powerful generation capability of LLMs, incorporating a retrieval module to retrieve relevant information from large-scale corpora demonstrates excellent superiority in the question answering (QA) and dialogue generation tasks \cite{RAG}.
						
						The broad application of data augmentation techniques effectively alleviates data scarcity issues and promotes the breakthrough and development of LLMs in diverse NLP tasks. Table \ref{DA task, sub-task, dataset} shows the popular tasks of data augmentation in recent research.

\subsection{Evaluation Metrics}
						
						Automatic and human evaluation are two main perspectives on a model's performance. Automatic evaluation is more suitable when the task contains a numerous dataset. Typical automatic evaluation includes accuracy \cite{WANLI,LLM2LLM,FlipDA}, exact match score \cite{Generative-DA,MRC-QA} and F1 Score \cite{LLM-DA,X-GEAR,Read-Com}.
						
						In contrast to automatic evaluation, human evaluation could deeply analyse the model's output through the involvement of crowd workers or experts. Common human evaluation criteria include consistency \cite{AugESC,UniMS-RAG,CGRG,SeeKeR}, coherence \cite{UniMS-RAG,ChatPLUG} and informativeness \cite{AugESC,ChatPLUG}. Human evaluation could provide a more comprehensive understanding of the model's performance, but it could be time-consuming and labour-intensive on massive datasets. The detailed evaluation of data augmentation studies is illustrated in table \ref{Evaluation}.

%% file: sec-Challenge.tex
Although data augmentation has showcased excellent performance in many studies and tasks, it still faces challenges and limitations. This chapter underscores some directions that could be investigated in greater depth for future research.

\subsection{The Quality and Diversity of Generated Data}

The prerequisite for effective data augmentation is the validity of the generated data. Even though generating semantically rich data that is highly relevant to seed training data \cite{HiPSTG,PROMPTING,LLM-powered} and employing various filtering strategies to filter out repetitive and irrelevant content generated by LLMs could address the limitations of the low-quality augmented datasets \cite{G-DAUGc,LLM-DA,RGQA}. However, there is no exact solution to determine how much synthetic data is efficient and how to ensure that the synthesised data is helpful to improve the performance of the model \cite{Promptagator}.

\subsection{Tasks Adaptation}

Current experiments mainly focused on a single task, such as classification tasks. Inputs and test inputs need more flexible options for sharing between diverse tasks \cite{zicl}. Specifically, the issues of adapting the model to different tasks and correctly evaluating the model's output on distinct tasks are unresolved \cite{EAE-RAG}.

\subsection{Reducing Hallucination}

LLMs' transformative generative ability has brought enormous benefits to both academic and industrial, but LLMs unavoidably produce factually incorrect responses and contradictory content \cite{SeeKeR}. Many studies propose to mitigate the hallucination by retrieving external and up-to-date knowledge and improving the accuracy of the generated content by retrieving relevant articles and providing citations for LLMs \cite{ALCE}. Another popular approach is to incorporate grounded truths through multiple APIs \cite{ReAct,SeeKeR,ChatPLUG}, search engines \cite{DialogGen}, and modulars \cite{SeeKeR}, thereby alleviating hallucination. 

\subsection{Retrieval Dependencies}
During the retrieval process, the model's performance depends to some extent on the quality of the retrieved data and the relevance between external and existing data \cite{RADA}. If the retriever fails to retrieve related grounded facts, the model may produce unfaithful answers \cite{KAPING,RADA}.

\subsection{Large Number of Parameters and High Training Cost}

LLMs have numerous training parameters, which take up a significant amount of GPU resources during the training process. The inference procedure also requires high computational and storage costs \cite{RGQA,RADA}. Future work could explore innovative methods to better transfer knowledge from large-scale language models to smaller language models. Investigate how to effectively integrate various domain knowledge into the model to improve its adaptability to complex tasks rather than increasing the model parameters.

\subsection{Ethics and Potential Risks}

LLMs may pose uncontrolled risks \cite{CGRG}, as their synthesised content may contain sensitive and private information \cite{ConvAug}. In addition, LLMs have inherited biases regarding specific topics, which may produce harmful content \cite{UDAPDR}.

Applying text data augmentation techniques in large language models still faces unresolved challenges. Future research could focus on optimising the quality of the generated data, improving the model's adaptability to different tasks, and reducing the training cost to further promote the effectiveness and development of data augmentation.

%% file: sec-CONCLUSION.tex
This survey classifies text data augmentation into four techniques: Simple Augmentation, Prompt-based Augmentation, Retrieval-based Augmentation, and Hybrid Augmentation. Each technique is further categorised according to its unique characteristics. Providing crafted prompt templates to large language models has shown prominent performance in data augmentation. Combining external data with existing data through a retriever offers more possibilities for cross-domain tasks. Data augmentation indeed contributes to expanding the dataset and generating diverse content, but more techniques are needed to test the validity and factuality of the generated data. The continuous development and improvement of LLMs makes them increasingly effective in data augmentation and deserves continued exploration in the future. 